\documentclass[a4paper,fleqn]{cas-dc}

\usepackage[authoryear,longnamesfirst]{natbib}
\usepackage{stfloats}
\usepackage{subfig}
\usepackage{balance}
\usepackage{algorithmic}
\usepackage{algorithm}
\usepackage{soul,color,xcolor}

\def\tsc#1{\csdef{#1}{\textsc{\lowercase{#1}}\xspace}}
\tsc{WGM}
\tsc{QE}
\tsc{EP}
\tsc{PMS}
\tsc{BEC}
\tsc{DE}

\begin{document}
	\let\WriteBookmarks\relax
	\def\floatpagepagefraction{1}
	\def\textpagefraction{.001}
	\shorttitle{}
	\shortauthors{Z. Yin et~al.}
	
	\title [mode = title]{Dynamic Programming-Based Offline Redundancy Resolution of Redundant Manipulators Along Prescribed Paths with Real-Time Adjustment}

	\author[1]{Zhihang Yin}[style=chinese]
	\fnmark[1]

	\affiliation[1]{organization={School of Mathematical Sciences, Zhejiang University},
		addressline={Xihu}, 
		city={Hangzhou},
		postcode={310058}, 
		state={Zhejiang},
		country={China}}
	
	\author[2]{Fa Wu}[style=chinese]
	\fnmark[1]

	\affiliation[2]{organization={Zhejiang Demetics Medical Technology Co., Ltd.},
		addressline={Xihu}, 
		city={Hangzhou},
		postcode={310012}, 
		state={Zhejiang},
		country={China}}
	
	\author[3]{Ziqian Wang}[style=chinese]

	\affiliation[3]{organization={Department of Human Development, Teachers College, Columbia University}, 
		city={New York},
		postcode={10027}, 
		state={NY},
		country={United States}}
	
	\author[4]{Jianmin Yang}[style=chinese]

	\affiliation[4]{organization={Zhejiang College of Sports},
		city={Hangzhou},
		postcode={311231}, 
		state={Zhejiang},
		country={China}}
	
	\author[5]{Jiyong Tan}[style=chinese, orcid=0000-0001-6356-1743]
	\cormark[1]
	\ead{scutjy2015@163.com}
	
	\affiliation[5]{organization={Shenzhen Institute for Advanced Study, University of Electronic Science and Technology of China}, 
		city={Shenzhen},
		postcode={610056}, 
		state={Guangdong},
		country={China}}
	
	\author[1]{Dexing Kong}[style=chinese, orcid=0000-0001-9339-8086]
	\cormark[1]
	\ead{dxkong@zju.edu.cn}

	\cortext[cor1]{Corresponding author}
	\fntext[fn1]{The two authors contribute equally to this work.}

	\begin{abstract}
		Traditional offline redundancy resolution of trajectories for redundant manipulators involves computing inverse kinematic solutions for Cartesian space paths, constraining the manipulator to a fixed path without real-time adjustments. Online control algorithms can achieve real-time adjustment of paths, but it cannot consider subsequent path points, leading to the possibility of the manipulator being forced to stop mid-motion due to joint constraints. To address this, this paper introduces a dynamic programming-based offline redundancy resolution for redundant manipulators along prescribed paths with real-time adjustment. The proposed method allows the manipulator to move along a prescribed path while implementing real-time adjustment along the normal to the path. Using Dynamic Programming, the proposed approach computes a global maximum for the variation of adjustment parameters. As long as the variation of adjustment parameters between adjacent sampling path points does not exceed this limit, the algorithm provides the next path point’s joint angles based on the current joint angles, enabling the end-effector to achieve the adjusted Cartesian pose. The main innovation of this paper lies in augmenting traditional offline optimal planning with real-time adjustment capabilities, achieving a fusion of offline planning and online control.
	\end{abstract}


	\maketitle
	
	\section{Introduction}
	
	In many applications of robotics, such as ultrasound scanning robots \citep{ref1}, massage robots \citep{ref0}, polishing robots \citep{ref01,refn3}, and others, it is necessary for the robot to move precisely along a pre-planned path in a three-dimensional workspace. When performing these tasks, non-redundant manipulators are easily limited by singular points and joint ranges, making it difficult for them to reach certain specific poses along the given path, so redundant manipulators are preferred{ \citep{ref3}}.
	
	In this paper, we consider the application scenario of robotic ultrasound scanning using a 7-degree-of-freedom (7-DOF) redundant manipulator. Equipped with an ultrasound probe at its end effector, the manipulator acquires ultrasound images in close proximity to the patient's skin. Consequently, the manipulator needs to move along a Cartesian space path specified according to the patient's body surface, aiming to complete the scan as seamlessly as possible to enhance the quality and efficiency of the diagnose. 
	
	{Ultrasound scanning necessitates the control of the contact force between the probe and the patient: excessive contact force may cause discomfort to the patient, while insufficient contact force can result in unclear images. Therefore, it is essential for the manipulator to actively adjust the contact force in real-time during the scanning process. Since the probe is connected to the end-effector of the manipulator by a spring, by adjusting the manipulator path along the direction of the probe, the contact force can be controlled by altering the length of the spring.}
	
	Due to the redundancy of the manipulator we use, a given end-effector pose can correspond to an infinite number of possible joint configurations, known as its inverse kinematic solutions. Although these joint configurations can all position the end-effector at the same pose, the range of joint configurations achievable during subsequent motion will differ due to constraints on joint angles, velocities, and other factors, leading to variations in the range of poses that can be later reached. Therefore, selecting the optimal solution from the infinite possible inverse kinematic solutions is crucial for the completeness of subsequent motion. This pertains to the redundancy resolution problem in redundant manipulators. The redundancy resolution algorithms for manipulators can be broadly categorized into two types: {online control} and offline planning.
	
	{In online control algorithms, the inverse kinematic solution of the target pose are computed based on the current state of the manipulator without regard to future states.} This method only considers the local path point, disregarding the restrictions of the current joint positions on subsequent motion, and thus cannot guarantee complete motion along a given path. The Cartesian pose controller integrated within the manipulator that we utilize belongs to this category. We will demonstrate its disadvantage through experiments in Section 4.1. On the other hand, {online control algorithms} inherently consider only the current state of the manipulator and the target pose, allowing for easy path adjustments. Widely used impedance control \citep{refn5,refn4,refn6,refn7} falls into this category of algorithms.
	
	To prevent the manipulator from halting its motion due to joint limitations during operation, a viable approach is to employ an offline planning algorithm. Such algorithms consider all the path points along the manipulator's path simultaneously before its motion, thoroughly addressing the restrictions imposed by the joint angles on subsequent motion. By considering all path points, the obtained inverse kinematic solutions can satisfy various joint constraints, such as those related to angle and velocity. However, due to the nonlinearity of the inverse kinematic mapping of the manipulator, changing the pose of a single path point and calculating its corresponding joint angles does not allow for a simple linear calculation of the subsequent points' corresponding joint angles. {Instead, the inverse kinematic solutions of all subsequent Cartesian poses need to be recomputed using the time-consuming algorithm, making real-time path adjustments challenging.}
	
	We aspire for the manipulator to traverse a path without interruptions while also being capable of real-time path adjustments during its motion. Therefore, we aim to enhance the offline planning algorithm to integrate real-time adjustment functionalities. Driven by this concept, we propose a dynamic programming-based offline optimal planning algorithm of redundant manipulators along prescribed paths with real-time adjustment. With this algorithm, the manipulator can complete its motion along the path while allowing real-time adjustments to the path.
	
	\subsection{Related Work}
	The key to our study lies in solving the redundancy resolution of a redundant manipulator. Redundancy resolution can mainly be done in the velocity or position level\citep{ref3F}. Algorithms in acceleration or higher-order level are also possible \citep{refr1}. Algorithms designed on the joint velocity or higher level involve online computing the pseudoinverse of the Jacobian matrix associated with the forward kinematics of the controlled manipulator\citep{ref4}. Such algorithm is online, as the manipulator moves, it calculates the joint angles for the next time step based on the current joint parameters and the expected Cartesian pose for the next time step. In \citep{ref4.1}, a real-time optimization problem is established by considering the joint constraints of the manipulator as constraints, with the error between the current trajectory of the manipulator and the expected trajectory as the optimization objective. RNN is used for real-time optimization.  
	
	However, as mentioned before, such velocity-based algorithm overlooks the influence of the current manipulator joint angles on the subsequent motion of the robot. As the manipulator moves, the joint angles of some joints may reach limits, causing the manipulator to jam and preventing further motion. Many algorithms have been proposed for joint-limit and singularity avoidance in online motion. The algorithm proposed in \citep{ref4.4} keeps the manipulator closer to the mid-joint position of the joint limits to eliminate the danger of joint limits. However, due to the uncertainty of subsequent paths, this still does not guarantee that the joint angles of the manipulator will not reach their limits. Similarly, \cite{refr4} maintains the manipulator as close as possible to the mid-joint position and as far as possible to the singularities at the same time to avoid joint limits and kinematic singularity. The algorithm proposed in \citep{refr2} devides tasks into different priority sets and allows a general number of scalar set-based tasks to be handled with a given priority within a number of equality tasks. This method can fulfill the system’s equality tasks while ensuring that the set-based tasks are always satisfied. With their algorithm, high-priority set-based tasks remain in their valid set at all times, whereas lower-priority set-based tasks cannot be guaranteed to be satisfied due to the influence of the higher-priority equality tasks. \cite{refr3} tested the set-based task-priority kinematic control scheme for a dual-arm mobile manipulator, ensuring the integrity of the system and the effectiveness of the mission to be accomplished. However, due to the nature of online motion generation, these methods cannot ensure that the manipulator moves along the prescribed path while simultaneously avoiding joint limits. 
	
	Algorithms designed on the joint position level generally introduce appropriate parameters to obtain the inverse kinematic solutions at various points along the path. Subsequently, by designing a global energy function, an optimization problem is formulated to determine the parameter values at each point, thereby achieving the global optimal solution for redundancy resolution along the path. Due to the simultaneous consideration of all path points and the computation of all inverse kinematic solutions, this type of algorithm belongs to offline planning and needs to be completed before the manipulator's motion. \citep{refr5} uses the joint space decomposition and selects redundant variables from the joint position vector as the redundancy parameters, obtaining a closed-form of inverse kinematic solution. Other parameterization methods, such as those proposed in \citep{ref6}, can also be selected based on the specific structure of the manipulator. 
	
	Through parameterization, determining the inverse kinematic solution for the manipulator's path transforms into finding the parameter values for each path point. This converts the problem into an optimization problem, where the objective of the optimization is determined by the practical requirements of the task. The algorithms proposed in \citep{refr5, ref6.5, ref7} use dynamic programming for optimal planning of redundant robots along prescribed paths. Based on the specific requirements of different practical tasks, an arbitrary energy function can be established as the objective of the optimization problem. Constraints can also be converted into constraints for the optimization problem through mathematical modeling. Since these algorithms calculate the joint angles corresponding to each moment of the manipulator's motion in advance, they can ensure the completeness of the manipulator's path. However, this also prevents the manipulator from making real-time adjustments to the path.
	
	Dynamic programming is a mathematical method for solving a class of optimization problems. It breaks down complex problems into a series of interconnected and overlapping subproblems, and combines the solutions of these subproblems recursively to obtain the optimal solution to the original problem\citep{ref8,ref9}. Using the dynamic programming algorithm, we can determine the globally optimal solution for redundancy resolution corresponding to the prescribed path, ensuring that the joint angle motion at every moment complies with the manipulator's constraints, {obtaining feasible trajectory results for the manipulator}. 
	
	To ensure the integrity of the manipulator's motion, this paper opts to use the dynamic programming-based offline planning algorithm as the foundation, combined with a real-time control algorithm to address the problem proposed in this study.
	
	\subsection{Paper Contribution and Organization}
	
	This paper's contribution lies in its pioneering integration of real-time path adjustment functionality into traditional offline planning algorithms. Conventional offline planning algorithms precompute the complete path of the manipulator before operation, ensuring the feasibility of the manipulator's motion but restricting it to follow the predetermined path. The proposed algorithm not only retains the advantages of offline planning by guaranteeing the feasibility of the manipulator's motion, but also enables real-time adjustment of the subsequent path based on feedback from sensors and other sources during operation, thereby achieving online control. The proposed algorithm represents a significant breakthrough in offline planning algorithms for manipulators. This paper presents three main innovations: 
	
	$\bullet$ Augmenting traditional offline optimal planning with real-time adjustment capabilities achieves a fusion of offline planning and online control, enabling the manipulator to adapt instantly to changes in the working environment and execute pre-defined tasks more effectively.
	
	$\bullet$ For the same pose, different inverse kinematic solutions are chosen based on the current joint angles, ensuring that the motion of each joint complies with the constraints.
	
	$\bullet$ The design of the dynamic programming algorithm enables the simultaneous determination of the inverse kinematic solution for the path and the maximum variation in path adjustment parameters.
	
	This paper is organized as follows. First, the mathematical model of the problem is described in section II. Then, the dynamic programming-based offline optimal planning algorithm of redundant manipulators along prescribed paths with real-time adjustment is proposed in section III. Finally, we tested the algorithm using test paths and randomly generated adjustment parameters, thereby demonstrating the reliability of the algorithm in section IV.
	
	\section{Preliminaries and problem formulation}
	
	In our work, we will take the Franka 7-DOF manipulator\citep{refn1} for study and experimentation. The host system communicates with the manipulator at a frequency of 1000Hz. The host system is capable of real-time access to various motion parameters of the manipulator, enabling it to send control commands for the motion of the manipulator. Therefore, we choose to discretize the pre-planned path in Cartesian space, ensuring that the sampling interval between adjacent path points is a multiple of the communication period of the manipulator.
	
	For a given path, we refer to the points that specify the corresponding Cartesian poses as sampling points, and the points at which the manipulator needs to communicate with the host and receive control signals as communication points. A sampling point necessarily corresponds to a communication point, while a communication point may not be a sampling point. The host system, based on real-time information obtained from sensor readings, adjusts the positions of subsequent path points along the normal to the path at sampling points.
	
	Given a discrete path $\{\mathbf{T}_{\text{EE}n}\}$ with $n+1$ sampling points in Cartesian space, $ i $ represents the index of sampling points ranging from 0 to $n$ and $\mathbf{T}_{\text{EE}i}$ denotes the end effector pose of the $ i $-th sampling point. $\mathbf{T}_{\text{EE}i}$ is a homogeneous transformation matrix, which can be represented by a three-dimensional rotation matrix and a three-dimensional translation vector:
	\begin{equation}
		\mathbf{T}_{\text{EE}i}=\begin{bmatrix}
			\mathbf{R}_{\text{EE}i}	&\mathbf{p}_{\text{EE}i} \\
			\mathbf{0}_{1\times 3}	&1
		\end{bmatrix}
	\end{equation}
	where $\mathbf{R}_{\text{EE}i}$ is the rotation matrix of the end effector of the $ i $th sampling point, and $\mathbf{p}_{\text{EE}i}$ is the translation vector. {Let $\mathbf{Z}_i$ denote the z-axis of the rotation matrix representing the end-effector orientation of the manipulator at the  $i$-th path point. This vector signifies the direction of the manipulator's end-effector. For instance, when the manipulator is employed for ultrasound scanning, $\mathbf{Z}_i$ corresponds to the direction of the ultrasound probe. }During the operation of the manipulator, the host will calculate the bounded adjustment parameter $y_i$ in real time and adjust the end effector pose of the $ i $th sampling point to $\mathbf{\hat{T}}_{\text{EE}i}(y_i)$ according to $\mathbf{Z}_i$:
	\begin{equation}
		\mathbf{\hat{T}}_{\text{EE}i}(y_i)=\begin{bmatrix}
			\mathbf{R}_{\text{EE}i}	&\mathbf{p}_{\text{EE}i}+y_i \mathbf{Z}_i \\
			\mathbf{0}_{1\times 3}	&1
		\end{bmatrix}
	\end{equation}
	
	We will only consider the limitations on joint angle and velocity at first. Constraints on acceleration and jerk can be achieved through velocity limitations and further algorithmic adjustments. The angular velocity of the manipulator's joints is limited by $\dot{\mathbf{q}}_{\text{max}}$, and the joint angles have both lower and upper bounds, denoted as $\mathbf{q}_{\min}$ and $\mathbf{q}_{\max}$ respectively. 
	
	Due to limitations in the angular velocity of the manipulator’s joints, the manipulator may fail to reach the adjusted position when the adjustment parameters undergo significant changes. {Therefore, we have to set a consistent upper bound $\delta$ on the change in the adjustment parameter $y_i$. As long as the difference between the adjustment parameter $ y_{i+1}$ of the next waypoint and $y_i$ of the current waypoint does not exceed this global maximum variation, the algorithm can compute a feasible inverse kinematic solution of $\mathbf{\hat{T}}_{\text{EE}i+1}(y_{i+1})$ in real-time. The upper bound  should be as large as possible to enable the manipulator to adjust the path to the maximum extent.}
	
	In offline planning, the assurance that the manipulator's trajectory complies with the joint constraints is established by predetermining the joint angles of each path point before the motion, ensuring that these angles satisfy the constraints. Building upon this concept, we want to calculate the joint angles corresponding to all path points in advance. Discretization is standard approach to offline redundancy resolution problems. Therefore, we divide the range of $y_i$ into 2$o$ equal parts and discretize the coefficients to $2o+1$ discrete values $\{b_{-o}, b_{-o+1}, \ldots, b_o\}$ to maintain a finite set of potential path scenarios.{ Given that the values of $y_i$ have already been uniformly discretized, for convenience, we can directly use the change in the indices of $y_i$ values in the sequence $\{b_{-o}, b_{-o+1}, \ldots, b_o\}$ to measure the variation in $y_i$. Denote $c_i$ as the indice of $y_i$ in the potential value sequence, which means $y_i=b_{c_i}$, and the restriction $\delta$ on the variation of $y_i$ implies a restriction on the variation of $c_i$, denoted as $d$. In other words, depending on the different values of $y_i$, $y_{i+1}$ can have at most $2d+1$ different values.}  
	
	Taking into account the constraints imposed by the joint angles of the previous sampling point on the current sampling point, for different adjusted paths, despite the manipulator having the same Cartesian pose at the current sampling point, the pose and joint angles of the previous sampling point may differ, leading to different constraints on the joint angles at the current sampling point. Thus, even with identical Cartesian poses, the corresponding joint angles differ. We will compute the inverse kinematic solution for the target pose based on the joint angles of the previous sampling point, in other words, obtaining the inverse kinematic mapping $\mathbf{q}_i = f^{-1}(\mathbf{\hat{T}}_{\text{EE}i}(y_i),\mathbf{q}_{i-1})$. {The previous joint angles $\mathbf{q}_{i-1}$ will serve as the redundancy parameter.}
	
	Deriving the inverse kinematic solutions based on the current joint angles of the manipulator resembles online planning. The distinction lies in the fact that we do not compute the inverse kinematic solutions in real-time, but rather perform all calculations before the manipulator's motion commences, ensuring uninterrupted motion of the manipulator under all feasible adjusted paths. 
	
	The problem we aim to address is as follows: to find the theoretical global maximum value for $d$, along with an inverse kinematic mapping $f^{-1}$, such that: for any $\{c_n\}$ that satisfies:
	
	\begin{equation}
		\begin{aligned}
			&c_0=0\\
			&	|c_{i}-c_{i-1}| \leqslant d, \forall i \in \{1,2,...,n\}
		\end{aligned}
	\end{equation}
	
	$\mathbf{q}_i = f^{-1}(\mathbf{\hat{T}}_{\text{EE}i}(b_{c_i}),\mathbf{q}_{i-1})$ is an inverse kinematic solution of $\mathbf{\hat{T}}_{\text{EE}i}(b_{c_i}) $, and $\mathbf{q}_i $ satisfies the following equation:
	
	\begin{equation}
		\begin{aligned}
			{q}_{\min,c} \leqslant q_{i,c} \leqslant {q}_{\max,c}, i = 0,1,...,n; c = 1,2,...,7 \\
			|\dot{q}_{i,c}|\leqslant {\dot{q}}_{\max,c}, i = 1,2,...,n; c = 1,2,...,7
		\end{aligned}
	\end{equation}
	
	where ${q}_{\min,c}, {q}_{\max,c}, q_{i,c}, \dot{q}_{i,c}, \dot{q}_{\max,c},  $ represent the $c$-th components of vectors $\mathbf{q}_{\min},\mathbf{q}_{\max}, \mathbf{q}_i, \dot{\mathbf{q}}_i, \dot{\mathbf{q}}_{\max} $ respectively. The angular volocities of joints can be easily calculated by definition:
	\begin{equation}
		\begin{aligned}
			\dot{\mathbf{q}}_{i} = \frac{\mathbf{q}_{i}-\mathbf{q}_{i-1}}{t_0}, i = 1,2,...,n
		\end{aligned}
	\end{equation}
	
	The process of the proposed algorithm is roughly illustrated as in Figure 1. The algorithm is roughly divided into two parts: the offline planning before the manipulator's motion and the real-time interaction with the manipulator during its motion, including the real-time adjustment of the path and motion compensation between adjacent sampling points.
	
	\begin{figure}[h]
		\centering
		\includegraphics[width=3.5in]{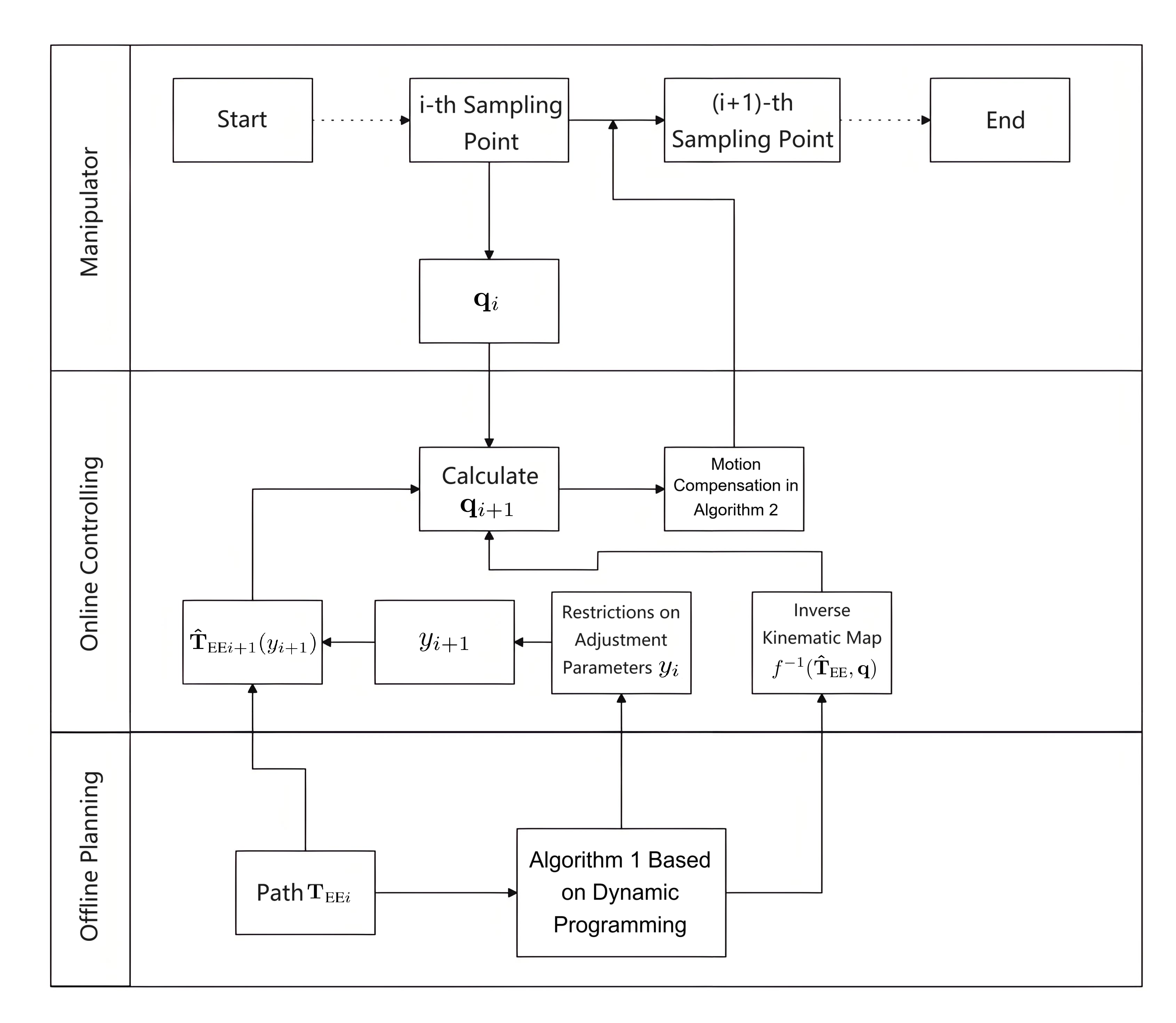}
		\caption{The process of the proposed algorithm. The algorithm is roughly divided into two parts: the offline planning before the manipulator's motion and the real-time interaction with the manipulator during its motion, including the real-time adjustment of the path and motion compensation between adjacent sampling points.} 
		
		\label{fig_0}
	\end{figure}

	\section{Algorithms}
	
	In this section, we will use dynamic programming to obtain the maximum $d$ and the inverse kinematic mapping $f^{-1}$. Subsequently, we will employ motion compensation algorithm to ensure that the joint trajectory satisfies constraints on acceleration and jerk, enabling it to operate on the actual hardware.
	
	\subsection{Inverse Kinematics}
	
	Before seeking the redundancy resolution for the entire path, it is worthwhile to first consider the inverse kinematic solution corresponding to a single Cartesian pose $\mathbf{T}_\text{EE}$. Due to the redundancy of the 7-DOF manipulator, there exist infinitely many inverse kinematic solutions. 
	
	Take the Franka Emika manipulator we use for experimental purposes in this study as an example. It is a nimble lightweight robot with 7 degrees of freedom and highly sensitive sensors. Figure 2 shows the Denavit-Hartenberg parameters of the Franka Emika manipulator, and Table 1 shows the limits of the joints of the manipulator\footnote{https://frankaemika.github.io/docs/}.
	
	\begin{figure}[h]
		\centering
		\includegraphics[width=3.5in]{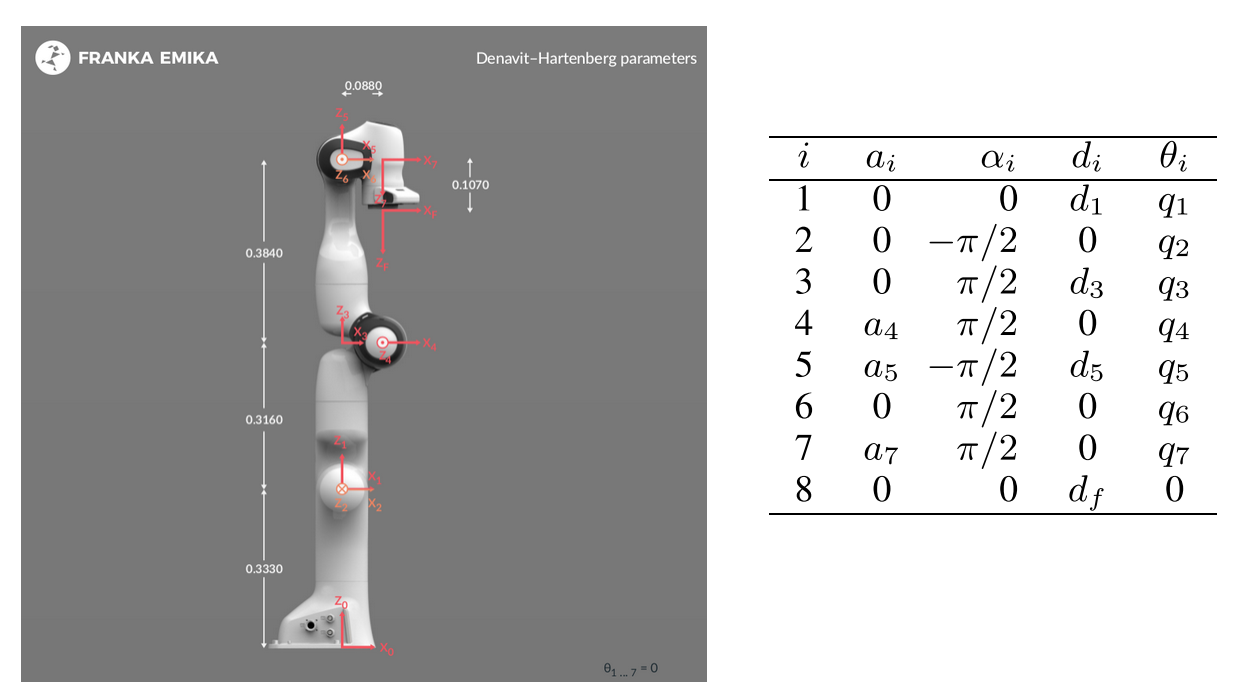}
		\caption{Denavit-Hartenberg frames and table of parameters for the Franka Emika robot. In this figure, joint angles $ q_{1,...,7} = 0 $. The reference frames follow the modified Denavit-Hartenberg convention, $ d_1 = 0.333 \text{m}, d_3 = 0.316 \text{m}, d_5 = 0.384 \text{m}, d_f = 0.107 \text{m}, a_4 = 0.0825 \text{m}, a_5 = -0.0825 \text{m}, a_7 = 0.088\text{m}$.} 
		
		\label{fig_1}
	\end{figure}
	
	\begin{table*}
		\caption{Limits of Joints\label{tab:table1}}
		\centering
		\renewcommand{\arraystretch}{1.5}
		\begin{tabular}{c c c c c c c c c}
			\hline
			Name  &  Joint 1 & Joint 2  &  Joint 3  & Joint 4  &  Joint 5  &  Joint 6  &  Joint 7  &  Unit  \\
			\hline
			$ \mathbf{q}_{\max } $ & 2.8973 & 1.7628 & 2.8973 & -0.0698 & 2.8973 & 3.7525 & 2.8973 & rad \\
			$ \mathbf{q}_{\min } $ & -2.8973 & -1.7628 & -2.8973 & -3.0718 & -2.8973 & -0.0175 & -2.8973 & rad \\
			$ \dot{\mathbf{q}}_{\max } $ & 2.1750 & 2.1750 & 2.1750 & 2.1750 & 2.6100 & 2.6100 & 2.6100 & $ \frac{\text{rad}}{\text{s}} $ \\
			$ \ddot{\mathbf{q}}_{\max } $ & 15 & 7.5 & 10 & 12.5 & 15 & 20 & 20 & $ \frac{\text{rad}}{\text{s}^{2}} $ \\
			$ \dddot{\mathbf{q}}_{\max } $ & 7500&3750&5000&6250&7500&10000&10000 & $ \frac{\text{rad}}{\text{s}^{3}} $ \\
			\hline
		\end{tabular}
	\end{table*}

	By employing the parameterization technique\citep{refn1.5}  and geometrical analysis\citep{refn2}, we can derive closed-form inverse kinematics solutions for the manipulator. Following \citep{ref5}, a certain joint angle can be fixed as a parameter. The algorithm proposed in \citep{ref15} calculates the inverse kinematic solutions of the Franka robot by taking $q_7$, the joint angle of the 7th joint of the manipulator, as a parameter. Once $q_7$ is specified, the manipulator can be considered as a 6-degree-of-freedom one. When a solution exists, there are three possible scenarios that lead to multiple solutions, resulting in up to 8 sets of solutions. These multiple solutions can be mutually converted when they exist. By considering the constraints on each joint angle, an analysis is conducted to exclude the multiple solutions while preserving the existence of the solutions through the imposition of restrictions on the joint angles. From this, we can obtain a bijective inverse kinematic mapping $f^{-1}_{q_7}$ within the workspace of the 6-DOF manipulator:
	\begin{equation}
		[q_1, q_2, ..., q_6]^T = f^{-1}_{q_7}(\mathbf{T}_\text{EE})
	\end{equation}
	Thus, we have obtained the parameterized inverse kinematic solution for the 7-DOF manipulator:
	\begin{equation}
		\begin{aligned}
			\mathbf{q} & = \tilde{f}^{-1}(\mathbf{T}_\text{EE}, q_7)\\
			& = \begin{bmatrix}
				f^{-1}_{q_7}(\mathbf{T}_\text{EE})	\\
				q_7
			\end{bmatrix}
		\end{aligned}	
	\end{equation}

	During the computation of the inverse kinematic solutions, we also incorporate avoidance strategies for singularity. In robotics, the relationship between joint velocities and Cartesian velocities at the end-effector of a manipulator is commonly expressed using the Jacobian matrix as follows:
	\begin{equation}
		\mathbf{v}=J(\mathbf{q})\dot{\mathbf{q}}
	\end{equation}
	where $ \mathbf{v} $ represents the Cartesian velocity vector and $ J $ denotes the Jacobian matrix. When the Jacobian matrix becomes singular, the manipulator loses one or more degrees of freedom, resulting in restricted motion in a particular direction in Cartesian space regardless of the chosen joint velocities. This limitation is known as singularity in the robot's motion\citep{ref3}. 
	
	After obtaining a set of joint angles, we can compute the corresponding Jacobian matrix and examine whether the matrix is singular, thereby excluding singular configurations of the manipulator.

	In this paper, we will adopt another discretization approach: dividing the range of $q_7$ into $m-1$ equal parts, thereby restricting the values of $q_7$ to $m$ discrete values $\{a_1, a_2, \ldots, a_m\}$.
	
	For other redundant manipulators with potentially greater degrees of freedom, the method for solving the inverse kinematics remains the same, albeit requiring the introduction of additional parameters. 
	
	\subsection{Dynamic Programming-Based Algorithm}
	
	{We have proposed algorithm 1 to find the global maximum value of $d$ and the inverse kinematic mapping $f^{-1}$ defined in section 2.}
	
	It should be noted that given the end-effector pose $\mathbf{T}_\text{EE}$, not every $q_7$ value can yield a corresponding inverse kinematic solution due to various restrictions on the manipulator joint angles and mechanical structure. The following test path serves as an example:
	
	\begin{equation}
		\mathbf{{T}}_\text{EE}(t) = \begin{bmatrix}
			\cos(\theta(t))	& \sin(\theta(t)) & 0 &0.6+0.1\cos(\theta(t)) \\
			\sin(\theta(t))	& -\cos(\theta(t)) & 0 & 0.1\sin(\theta(t))\\
			0	& 0 & -1 & 0.1\\
			0	& 0 &0  &1
		\end{bmatrix}
	\end{equation}
	where $\theta(t) = \frac{2\pi}{ t_{\max}  }t - \pi, 0 \leqslant t \leqslant t_{\max}$ and the duration $t_{\max}$ is 10 seconds. {Sample this path at consistent time intervals to obtain  $n+1$  path points, and we can obtain the discretized path $\{\mathbf{T}_{\text{EE}n}\}$.}

	The existence of solutions corresponding to each $\mathbf{{T}}_\text{EE}(t)$ and $q_7$ is depicted in a figure, where the blue regions indicate the existence of the inverse kinematic solutions, as shown in Figure 3.

	\begin{figure}
		\centering
		\includegraphics[width=3.5in]{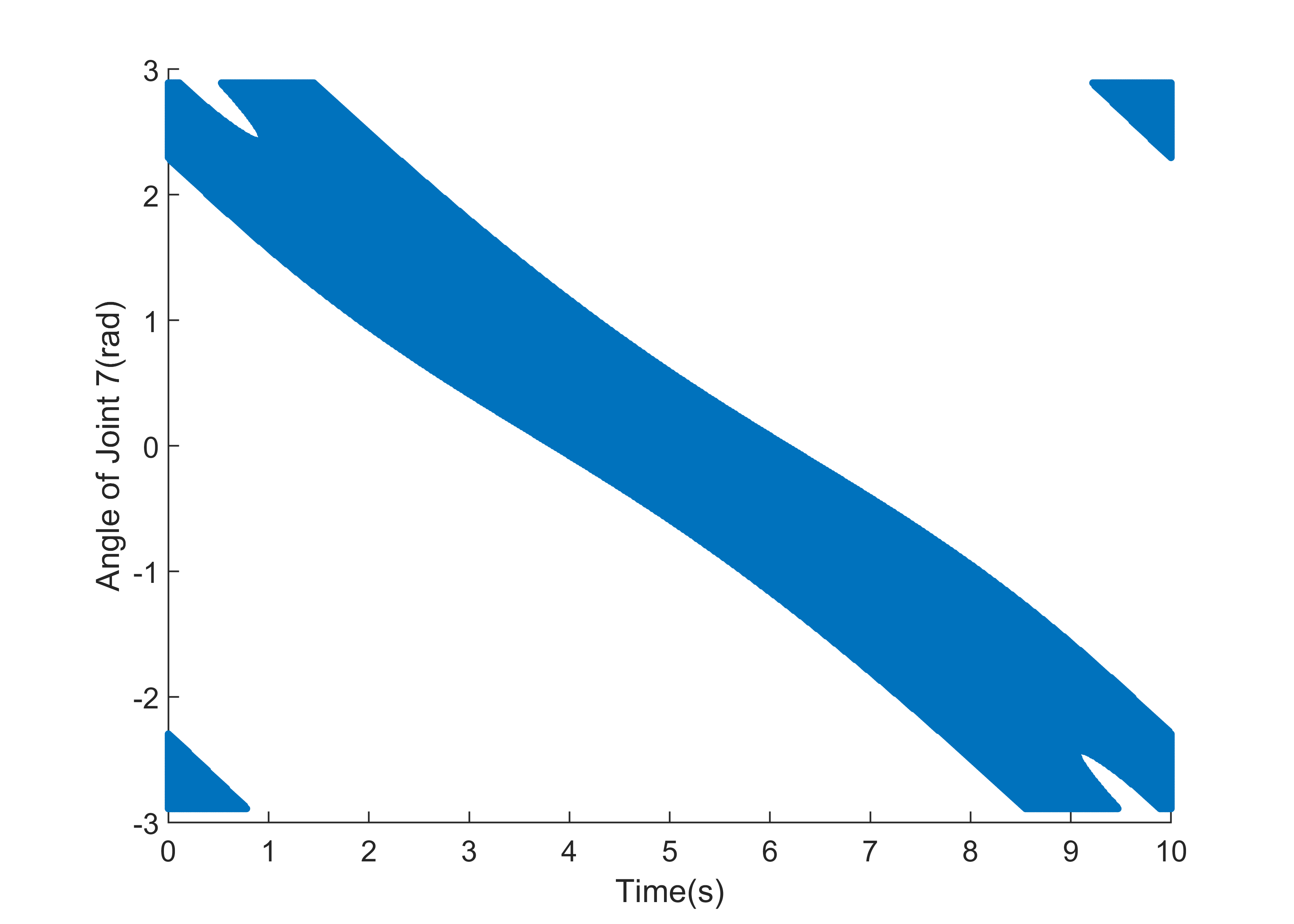}
		\caption{The existence of solutions corresponding to each $\mathbf{{T}}_\text{EE}(t)$ and $q_7$ is depicted in a figure, where the blue regions indicate the existence of the inverse kinematic solutions.} 
		
		\label{fig_2}
	\end{figure}
	
	For the inverse kinematic solution of a fixed path, it is tantamount to finding a continuous curve within the blue region in Figure 3 that connects $t=0$ and $t=10$, while ensuring that adjacent points on the curve satisfy the manipulator's constraints on joint velocities. Other joint angles, such as the point in the lower left blue section in Figure 3, although representing an inverse kinematic solution for the path point, are not contiguous with the subsequent blue sections, thus causing a discontinuity in motion. This underscores the necessity of anticipating subsequent path points in advance.
	
	The problem under investigation in this study introduces a adjustment parameter $y_i$ for the $i$th sampling point pose $\mathbf{T}_{\text{EE}i}$ in equation (2). Similarly, the adjusted pose $\mathbf{\tilde{T}}_\text{EE}(y)$ of a Cartesian pose $\mathbf{T}_{\text{EE}}$ corresponding to the adjustment parameter $y$ is defined as follows:
	
	\begin{equation}
		\mathbf{\tilde{T}}_{\text{EE}}(y)=\begin{bmatrix}
			\mathbf{R}_{\text{EE}}	&\mathbf{p}_{\text{EE}}+y \mathbf{Z} \\
			\mathbf{0}_{1\times 3}	&1
		\end{bmatrix}
	\end{equation}
	where $\mathbf{R}_{\text{EE}}$ is the rotation matrix of the end effector,  $\mathbf{p}_{\text{EE}}$ is the translation vector, and $\mathbf{Z}$ denote the z-axis of the rotation matrix $\mathbf{R}_{\text{EE}}$. We can depict the existence of the inverse kinematic solution for each $\mathbf{T}_\text{EE}(t)$ of the test path corresponding to $q_7$ and $y$ in a three-dimensional coordinate system. The colored region in the graph indicates the existence of the corresponding solutions, as illustrated in Figure 4.
	
	\begin{figure}
		\centering
		\includegraphics[width=3.5in]{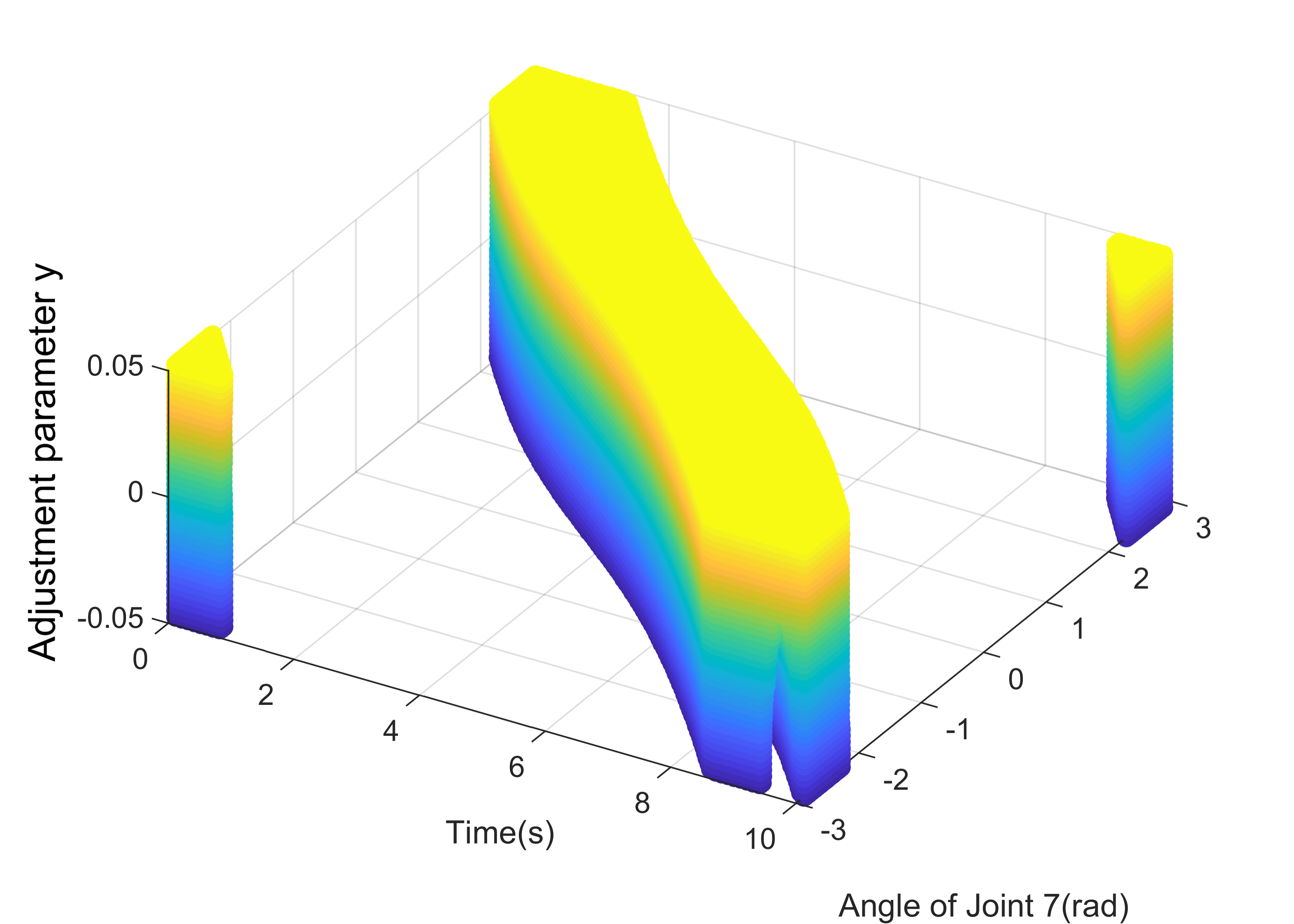}
		\caption{The existence of solutions corresponding to each $\mathbf{{T}}_\text{EE}(t)$, $y$ and $q_7$ is depicted in a figure, where the colored regions indicate the existence of the inverse kinematic solutions.} 
		
		\label{fig_3}
	\end{figure}
	
	Similar to the fixed path problem, we need to solve the inverse kinematics within the colored region of the three-dimensional coordinate system in Figure 4. Denote $ \tilde{f}^{-1}(\hat{\mathbf{T}}_{\text{EE}i}(b_k),a_j)$, {which is the inverse kinematic solution of the $i$th path point with the $j$th potential value $a_j$ of the 7th joint angle $q_7$ and the $k$th potential value $b_k$ of the adjustment parameter $y$,} as $\bar{\mathbf{q}}_{i,j,k}$. {The inverse kinematic solution $\tilde{f}^{-1}$ is defined in equation (7).}  The colored point with Cartesian coordinates $(\frac{t_\text{max}}{n}i,j,k)$ in Figure 4 represents $\bar{\mathbf{q}}_{i,j,k}$. We aim to find the global maximum variation $d$ between $c_i$ and $c_i+1$ for all $i$. To obtain this global maximum, we will employ a dynamic programming algorithm and construct a local energy function for each set of potential joint angles. 
	
	For the joint angles $\bar{\mathbf{q}}_{i,j,k} = f^{-1}(\hat{\mathbf{T}}_{\text{EE}i}(b_k),a_j)$ of the manipulator, we can consider this path point as a new starting point and move along the original path to the endpoint with adjustment, thus forming a sub-path. On this sub-path, we can define an energy function $L(i,j,k)$ analogous to $d$: $L(i,j,k)$ is the maximum $d_i$, such that for any $\{y_n\}$ that satisfies:
	
	\begin{equation}
		\begin{aligned}
			&	c_i = k, \mathbf{q}_i = \bar{\mathbf{q}}_{i,j,k}\\
			&	|c_{x}-c_{x-1}| \leqslant d_i, \forall x \in \{i+1,i+2,...,n\}
		\end{aligned}
	\end{equation}
	
	$\mathbf{q}_x = f^{-1}(\mathbf{\hat{T}}_{\text{EE}i}(b_{c_x}),\mathbf{q}_{x-1})$ is an inverse kinematic solution of $\mathbf{\hat{T}}_{\text{EE}x}(b_{c_x}) $, and $\mathbf{q}_x $ satisfies the following equation:
	
	\begin{equation}
		\begin{aligned}
			{q}_{\min,c} \leqslant q_{x,c} \leqslant {q}_{\max,c}, x = i,i+1,...,n; c = 1,2,...,7 \\
			|\dot{q}_{x,c}|\leqslant {\dot{q}}_{\max,c}, x = i+1,i+2,...,n; c = 1,2,...,7
		\end{aligned}
	\end{equation}
	{In the sub-path starting from the joint angles $\bar{\mathbf{q}}_{i,j,k}$, as long as the difference between the indice $ c_{x+1}$ of the next waypoint's adjustment parameter and $c_x$ of the current does not exceed $d_i$, the algorithm can compute a feasible inverse kinematic solution of $\mathbf{\hat{T}}_{\text{EE}x+1}(y_{x+1})$ in real-time.} It's obvious that :
	
	\begin{equation}
		\max d = \max_{j=1}^m {L}(0,j,0)
	\end{equation}
	
	{The sub-path starting from the joint angles $\bar{\mathbf{q}}_{i,j,k}$ can be further decomposed into the starting point and the sub-path starting from the next sampling point. We will utilize this decomposition to iteratively compute all values of $L(i,j,k)$. }The potential pose of the next sampling point is $\hat{\mathbf{T}}_{\text{EE}i+1}(b_{k+e})$, where $| e | \leqslant L(i,j,k)$, and $| k+e| \leqslant o$. For all $\bar{\mathbf{q}}_{i,j,k}$, we will recursively calculate $L(i,j,k)$ and define all inverse mappings $f^{-1}(\hat{\mathbf{T}}_{\text{EE}i+1}(b_{k+e}),\bar{\mathbf{q}}_{i,j,k})$. 
	
	For any chosen set of $i$, $j$, and $k$, and for any non-negative integer $s$, if $L(i,j,k) \geqslant s$, according to the definition, for any $\{c_n\}$ satisfying $|c_{x}-c_{x-1}| \leqslant s $ for all $ x = i+1,i+2,...,n$ and $c_i = k$, we have $\mathbf{q}_i = \bar{\mathbf{q}}_{i,j,k}$, $\mathbf{q}_x = f^{-1}(\mathbf{\hat{T}}_{\text{EE}x}(b_{c_x}),\mathbf{q}_{x-1})$ as an inverse kinematic solution of $\mathbf{\hat{T}}_{\text{EE}x}(b_{c_x}) $, and $\mathbf{q}_x $ satisfies the constraints. Take $c_{i+1}=k+e$ which satisfies $ |e|\leqslant s$ and $|k+e|\leqslant o$. $|c_{i+1}-c_{i}| \leqslant s$, thus there exists a $j_e$ such that $\bar{\mathbf{q}}_{i+1,j_e,k+e} = f^{-1}(\mathbf{\hat{T}}_{\text{EE}i+1}(b_{k+e}),\mathbf{q}_{i})$. $\bar{\mathbf{q}}_{i+1,j_e,k+e}$ is a pathpoint of the sub-path starting from $\bar{\mathbf{q}}_{i,j,k}$, thus for any $\{c_n\}$ satisfying $|c_{x}-c_{x-1}| \leqslant s $ for all $ x = i+2,i+3,...,n$ and $c_{i+1} = k+e$,  we can take $c_{i} = k$ and use the $\mathbf{q}_x$ calculated from the sub-path starting from $\bar{\mathbf{q}}_{i,j,k}$ as the inverse kinematic solution of $\mathbf{\hat{T}}_{\text{EE}x}(b_{c_x}) $ in the sub-path starting from $\bar{\mathbf{q}}_{i+1,j_e,k+e}$. It inherently satisfies the constraints. From definition, we have $L(i+1,j_e,k+e)\geqslant s$. Thus, if $L(i,j,k) \geqslant s$, for any $e$ satisfying $|e|\leqslant s$ and $|k+e|\leqslant o$, there exists a $j_e$ such that $L(i+1,j_e,k+e)\geqslant s$. 
	
	Conversely, if for any $e$ satisfying $|e|\leqslant s$ and $|k+e|\leqslant o$, there exists a $j_e$ such that $L(i+1,j_e,k+e)\geqslant s$ and $\bar{\mathbf{q}}_{i+1,j_e,k+e}$ satisfies the constraints with $\bar{\mathbf{q}}_{i,j,k}$, then $L(i,j,k) \geqslant s$. For any $\{c_n\}$ satisfying $|c_{x}-c_{x-1}| \leqslant s $ for all $ x = i+1,i+2,...,n$ and $c_{i} = k$,  there's a certain $e$ such that $c_{i+1} = k+e$. We can then use the $\mathbf{q}_x$ calculated from the sub-path starting from $\bar{\mathbf{q}}_{i+1,j_e,k+e}$ as the inverse kinematic solution of $\mathbf{\hat{T}}_{\text{EE}x}(y_x) $ in the sub-path starting from $\bar{\mathbf{q}}_{i,j,k}$. It inherently satisfies the constraints. From definition, we have $L(i,j,k) \geqslant s$. 
	
	Thus, we have obtained the recursive property of $L(i,j,k)$: $L(i,j,k)$ is the maximum $d_i$, such that for any $e$ satisfying $|e|\leqslant d_i$ and $|k+e|\leqslant o$, there exists a feasible $\bar{\mathbf{q}}_{i+1,j_e,k+e}$ satisfying joint angle and velocity constraints, and $L(i+1,j_e,k+e) \geqslant d_i$.
	
	For a fixed set of $i,j$ and $k$, we also need to define $f^{-1}(\hat{\mathbf{T}}_{\text{EE}i+1}(b_{k+e}), \bar{\mathbf{q}}_{i,j,k})$ for any $e$ satisfying $|e|\leqslant L(i,j,k)$ and $|k+e|\leqslant o$. We denote the inverse kinematic solution $\bar{\mathbf{q}}_{i+1,j',k+e}$ of $\hat{\mathbf{T}}_{\text{EE}i+1}(b_{k+e})$ that complies with the constraints and has the maximum $L(i+1,j',k+e)$ as $\bar{\mathbf{q}}_{i+1,j_e,k+e}$. We define: 
	\begin{equation}
		f^{-1}(\hat{\mathbf{T}}_{\text{EE}i+1}(b_{k+e}),\bar{\mathbf{q}}_{i,j,k})=\bar{\mathbf{q}}_{i+1,j_e,k+e}
	\end{equation}
	
	According to the recursive property, we will start from $i=n-1$ and recursively calculate all the values of $L(i,j,k)$ with $i$ moving backward. Algorithm 1 shows the solution:
	
	\begin{algorithm}
		\caption{The solution for the maximum value of $d$}\label{alg:alg1}
		\begin{algorithmic}[1]
			\STATE {\text{Set }}$i = n - 1$
			\STATE {\text{Set }}$d_{i} = 0$
			\STATE {For $1\leqslant j \leqslant m$, $-o\leqslant k \leqslant o$ }
			\STATE {Repeat:}
			\STATE \hspace{0.5cm}{\text{If feasible solutions $\bar{\mathbf{q}}_{i+1,j_{\pm d_{i}},k\pm d_{i}}$   both exist}} 
			\STATE \hspace{1.0cm}$d_{i} = d_{i} + 1$
			\STATE \hspace{0.5cm}{\text{Else, $L(i,j,k) = d_{i} - 1$, terminate the loop}}
			
			\STATE {\text{For }}$(n-1 > i \geqslant 0)$
			\STATE \hspace{0.5cm}{For $1\leqslant j \leqslant m$, $-o\leqslant k \leqslant o$ }
			\STATE {Repeat}
			\STATE \hspace{0.5cm}{\text{If feasible solutions $\bar{\mathbf{q}}_{i+1,j_{\pm d_{i}},k\pm d_{i}}$   both exist}} 
			\STATE \hspace{0.5cm}{And $L(i+1,j_t,k+t)\geqslant d_i$ , $\forall|t|\leqslant d_i$}
			\STATE \hspace{1.0cm}$d_{i} = d_{i} + 1$
			\STATE \hspace{0.5cm}{\text{Else, $L(i,j,k) = d_{i} - 1$, terminate the loop}}
			
			\STATE {\text{Find }}$j_0 $\text{, such that: }
			\STATE \hspace{0.5cm}${L}(0,j_0,0) = \max_{j=1}^m {L}(0,j,0)$ 
			\STATE \hspace{0.5cm}{Denote $\bar{\mathbf{q}}_{0,j_0,0}$ as the starting point of the path}
			\STATE \hspace{0.5cm}$d_{max} = {L}(0,j_0,0)$
		\end{algorithmic}
		\label{alg1}
	\end{algorithm}
	
	 For $\bar{\mathbf{q}}_{n-1,j,k}$, we simply need to ensure that the joint angles at the next sampling point satisfies the constraints. We start with $d_{n-1}$ at 0 and and increase it by 1 each time, checking if the compliant solutions $\bar{\mathbf{q}}_{i+1,j_{-d_{n-1}},k-d_{n-1}}$ and $\bar{\mathbf{q}}_{i+1,j_{d_{n-1}},k+d_{n-1}}$ both exist, until one does not. Then the previous $d_{n-1}$ yields the $L(n-1,j,k)$. Line 1 to Line 7 in Algorithm 1 correspond to this step.
	
	Assuming that for all $j'$ and $k'$, we have computed $L(i+1,j',k')$, we will use the computed $L(i+1,j',k')$ to recursively calculate $L(i,j,k)$. For any $\bar{\mathbf{q}}_{i,j,k}$, set $d_i=0$. Traversing $j'$ and check if $\bar{\mathbf{q}}_{i+1,j',k}$ satisfies constraints on joint angles and velocities. If there are no joint angles that satisfy the constraints, it indicates that the manipulator is stuck at the current joint angles and cannot proceed to the next step. Otherwise, there exists a $j_0$ such that $\bar{\mathbf{q}}_{i+1,j_0,k}$ satisfies the constraints and has the maximum $L(i+1,j_0,k)$. If $L(i+1,j_0,k)=0$, we have $L(i,j,k)=0$ and stop the increment on $d_i$.
	
	Then we move on to $d_i=1$. We traverse $\bar{\mathbf{q}}_{i+1,j',k-1}$ and $\bar{\mathbf{q}}_{i+1,j',k+1}$, and check if constraints on joint angles and velocities are satisfied. If there exist $j_{-1},j_1$ such that $\bar{\mathbf{q}}_{i+1,j_{-1},k-1}$ and $\bar{\mathbf{q}}_{i+1,j_1,k+1}$ satisfy the constraints respectively, and $L(i,j_{\pm 1}, k\pm 1)\geqslant 1$, by the recursive property, we have $L(i,j,k)\geqslant 1$. Otherwise, $L(i,j,k) = 0$.
	
	Subsequently, we increase on $d_i$ by $1$ at each step, traverse $\bar{\mathbf{q}}_{i+1,j',k-d_i}$ and $\bar{\mathbf{q}}_{i+1,j',k+d_i}$. If constraints cannot be satisfied, or any $L(i+1,j_{t},k+t) $ for $t\in \{ -d_i, -d_i+1,...,d_i\}$ is less than $d_i$, we have $L(i,j,k) = d_i-1$, and stop the increment on $d_i$. Line 8 to Line 14 in Algorithm 1 correspond to this step.
	
	Through the aforementioned recursion, we can compute all the values of $L(i,j,k)$. Subsequently, from $L(0,j,0)$, we select the maximum value $L(0,j_0,0)$, which represents the maximum $d$. The corresponding $\bar{\mathbf{q}}_{0,j_0,0}$ is then used as the inverse kinematic solution for the starting point of the path.  Line 15 to Line 18 in Algorithm 1 correspond to this step.
	
	During the motion of the manipulator, the $c_i$ corresponding to the $i$th sampling point is read in real-time. Suppose $\bar{\mathbf{q}}_{i-1,j,c_{i-1}} $ is the joint angles at the $(i-1)$-th sampling point, then $\bar{\mathbf{q}}_{i,j',c_i} = f^{-1}(\hat{\mathbf{T}}_{\text{EE}i}(b_{c_i}),\bar{\mathbf{q}}_{i,j,k})$ is used as the $i$-th sampling point's joint angle. This value is then sent to the manipulator, enabling it to move along the adjusted path.
	
	Based on the recursive structure and the number of iterations of the algorithm, we can derive its computational complexity. The complexity is related to the number of path points $n$, the number of possible values for the seventh joint angle $m$, and the number of adjustment parameter values $o$. The computational complexity of the algorithm is $O(m^2 n o^2)$.
	
	\subsection{Motion Compensation}
	
	In our study, we have only considered constraints on joint angle and velocity. However, in practical manipulator operations, there are also limitations on joint acceleration and jerk. Therefore, when testing the algorithm on the actual manipulators, adjustments to the algorithmic results are necessary to conform to these additional constraints. 
	
	We have chosen to employ a motion compensation algorithm. Due to the time intervals between sampling points along the path being greater than the communication cycle of the manipulator, the host system needs to communicate with the manipulator and issue control commands between adjacent sampling points. The target joint angles at intermediate points need to be obtained through interpolation or similar methods. To ensure smooth motion, we aim for the manipulator's joints to move at a constant speed between adjacent sampled points. Algorithm 2 is our motion compensation algorithm.
	
		\begin{algorithm}
		\caption{Motion Compensation}\label{alg:alg2}
		\begin{algorithmic}[1]
			\STATE {\text{Read the elapsed time $t$ of the motion}}
			\STATE {\text{If $t$ is the $i$-th sampling point}}
			\STATE \hspace{0.5cm} \text{Calculate $c_{i+1}$}
			\STATE \hspace{0.5cm} $\mathbf{q}_{i+1} = f^{-1}(\hat{\mathbf{T}}_{\text{EE}i+1}(b_{c_{i+1}}),\mathbf{q}_{i})$
			
			\STATE {\text{Calculate the remaining time $t_\text{r}$ before $\hat{\mathbf{T}}_{\text{EE}i+1}(b_{c_{i+1}})$}}
			\STATE {$\mathbf{\ddot{q}}_\text{max'} = \min(\mathbf{\ddot{q}}_\text{max}, \mathbf{\dddot{q}}_\text{max}\cdot n_0\cdot t_0)$}
			\STATE {$\mathbf{\dot{q}}_\text{max'} = \min(\mathbf{\dot{q}}_\text{max}, \mathbf{\ddot{q}}_\text{max}\cdot n_0\cdot t_0-\frac{\mathbf{\ddot{q}}_\text{max}^2}{2\cdot \mathbf{\dddot{q}}_\text{max}})$}
			\STATE {\text{Read $\mathbf{q}_\text{now}$,  $\mathbf{\dot{q}}_\text{now}$, and $\mathbf{\ddot{q}}_\text{now}$}}
			\STATE {$\mathbf{\dot{q}}_\text{d} = \frac{\mathbf{q}_{i+1}-\mathbf{q}_\text{now}}{t_\text{r}}$}
			\STATE {$\mathbf{\ddot{q}}_\text{d} = \frac{\mathbf{\dot{q}}_\text{d}-\mathbf{\dot{q}}_\text{now}}{t_0}$}
			\STATE {$\mathbf{\dddot{q}}_\text{d} = \frac{\mathbf{\ddot{q}}_\text{d}-\mathbf{\ddot{q}}_\text{now}}{t_0}$}
			\STATE {*For $1\leqslant c\leqslant 7$}
			\STATE \hspace{0.5cm} {If ${\dddot{q}}_{\text{d}c}> {\dddot{q}}_{\text{max}c}$ }
			\STATE \hspace{1cm} {${\dddot{q}}_{\text{d}c}= {\dddot{q}}_{\text{max}c}$, ${\ddot{q}}_{\text{d}c}={\ddot{q}}_{\text{now}c}+{\dddot{q}}_{\text{d}c}\cdot t_0,$ }
			\STATE \hspace{1cm} {${\dot{q}}_{\text{d}c}={\dot{q}}_{\text{now}c}+{\ddot{q}}_{\text{d}c}\cdot t_0$, return to *}
			\STATE \hspace{0.5cm} {If ${\ddot{q}}_{\text{d}c}> {\ddot{q}}_{\text{max'}c}$ }
			\STATE \hspace{1cm} {${\ddot{q}}_{\text{d}c}= {\ddot{q}}_{\text{max'}c}$, ${\dddot{q}}_{\text{d}c}=\frac{{\ddot{q}}_{\text{now}c}-{\ddot{q}}_{\text{d}c}}{t_0},$ }
			\STATE \hspace{1cm} {${\dot{q}}_{\text{d}c}={\dot{q}}_{\text{now}c}+{\ddot{q}}_{\text{d}c}\cdot t_0$, return to *}
			\STATE \hspace{0.5cm} {If ${\dot{q}}_{\text{d}c}> {\dot{q}}_{\text{max'}c}$}
			\STATE \hspace{1cm} {${\dot{q}}_{\text{d}c}= {\dot{q}}_{\text{max'}c}$, ${\ddot{q}}_{\text{d}c}=\frac{{\dot{q}}_{\text{now}c}-{\dot{q}}_{\text{d}c}}{t_0},$ }
			\STATE \hspace{1cm} {${\dddot{q}}_{\text{d}c}=\frac{{\ddot{q}}_{\text{now}c}-{\ddot{q}}_{\text{d}c}}{t_0}$, return to *}
			\STATE {$\mathbf{q}_\text{d} = \mathbf{q}_\text{now} + \mathbf{\dot{q}}\cdot t_0$}
			\STATE {Return $\mathbf{q}_\text{d}$ as the expected joint angle for the next communication point.}
			
		\end{algorithmic}
		\label{alg2}
	\end{algorithm}
	
	During communication between the host system and the manipulator, real-time readings of the current joint angles $\mathbf{q}_\text{now}$, velocities $\mathbf{\dot{q}}_\text{now}$, and accelerations $\mathbf{\ddot{q}}_\text{now}$ of joints are obtained. Adjustment coefficients read at the last sampling point and the algorithmically derived path are used to calculate the expected joint angles $\mathbf{q}_{i+1}$ at the next sampling point. Subsequently, the remaining time before the next sampling point, denoted as $t_\text{r}$, is used to compute the expected velocity $\mathbf{\dot{q}}_\text{d}$, acceleration $\mathbf{\ddot{q}}_\text{d}$ and jerk $\mathbf{\dddot{q}}_\text{d}$ for the next communication cycle:
	
	\begin{equation}
		\begin{aligned}
			\mathbf{\dot{q}}_\text{d}& = \frac{\mathbf{q}_{i+1}-\mathbf{q}_\text{now}}{t_\text{r}}\\
			\mathbf{\ddot{q}}_\text{d}& = \frac{\mathbf{\dot{q}}_\text{d}-\mathbf{\dot{q}}_\text{now}}{t_0}\\
			\mathbf{\dddot{q}}_\text{d}& = \frac{\mathbf{\ddot{q}}_\text{d}-\mathbf{\ddot{q}}_\text{now}}{t_0}
		\end{aligned}	
	\end{equation}
	where $t_0$ is the communication period of the manipulator. Lines 9 to 11 in Algorithm 2 correspond to this step.
	
	After that, we sequentially examine whether the jerk, acceleration, and velocity exceed their respective limits. If any of these values surpass the limits, they need to be adjusted to their corresponding maximum values, and the other two motion parameters need to be recalculated until all motion parameters comply with the constraints. With the calculated velocity, we can obtain the desired joint angles for the next communication point. Lines 12 to 23 correspond to this step.
	
	Considering the constraint on jerk, the manipulator may not be able to change the acceleration direction in a short time. After accelerating, if the manipulator needs to decelerate in a certain joint, there will still be a period of acceleration. This acceleration period might cause the joint angle to exceed its velocity limits. Therefore, a "cautionary value" needs to be set for the joint velocities. When the velocity surpasses this limit, the manipulator must reduce its acceleration and begin deceleration. 
	
	We can simulate the extreme scenarios for each joint, considering the need for the manipulator to immediately cease acceleration and begin deceleration when operating at maximum acceleration. At this point, the jerk reaches its upper limit and is opposite in direction to the current velocity. Assuming the velocity at the current communication point is 0, we can calculate the velocities and accelerations for the subsequent communication points. The velocity when the acceleration is 0 represents the velocity we need to reserve for the deceleration process. Subtracting this reserved velocity from the maximum velocity yields the cautionary value for velocity.
	
	Similarly, due to the constraint on manipulator acceleration, when the manipulator velocity needs to change direction, there will still be a short deceleration along the current direction. This deceleration may cause the joint angles to exceed their limits. Consequently, when using the algorithm to solve the inverse kinematics, the joint angle limits need to be reduced to accommodate space for deceleration.
	
	\begin{figure}
		\centering
		\includegraphics[width=3in]{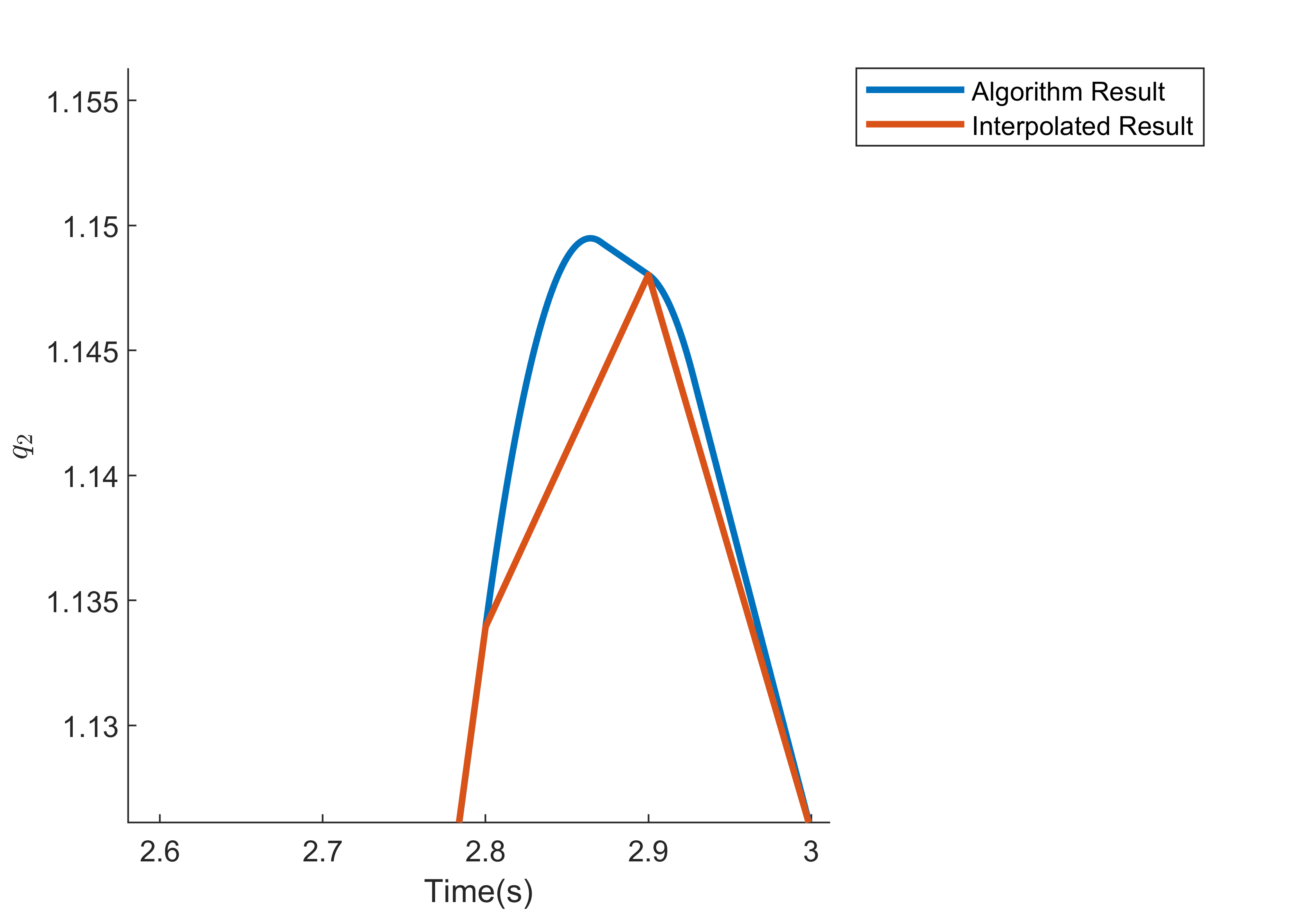}
		\caption{The interpolated expected path points cannot guarantee that acceleration and jerk constraints are satisfied, hence we can only modify the actual path under these constraints. This process inevitably leads to discrepancies between the actual and theoretical paths, and our algorithm minimizes this error to the greatest extent under joint constraints, ensuring that the manipulator smoothly reaches the expected joint angles at the next sampling point.} 
		
		\label{fig_4}
	\end{figure}
	
	Figure 5 succinctly illustrates our motion compensation algorithm. The interpolated expected path points cannot guarantee that acceleration and jerk constraints are satisfied, hence we can only modify the actual path under these constraints. This process inevitably leads to discrepancies between the actual and theoretical paths, and our algorithm minimizes this error to the greatest extent under joint constraints, ensuring that the manipulator smoothly reaches the expected joint angles at the next sampling point.
	
	In addition, we also need to consider how to smoothly bring the manipulator to a stop as the motion approaches its end. This requires the manipulator's joint velocity, acceleration, and jerk to be zero as the motion stops. In order to smoothly bring the manipulator's velocity and acceleration to zero, additional constraints need to be satisfied:
	\begin{equation}
		\begin{aligned}
			&\mathbf{\ddot{q}}\leqslant \mathbf{\dddot{q}}_\text{max}\cdot n_0\cdot t_0\\
			&\mathbf{\dot{q}}\leqslant \mathbf{\ddot{q}}_\text{max}\cdot n_0\cdot t_0-\frac{\mathbf{\ddot{q}}_\text{max}^2}{2\cdot \mathbf{\dddot{q}}_\text{max}}
		\end{aligned}
	\end{equation}
	where $n_0$ is the number of remaining communication cycles. Line 6 and line 7 in Algorithm 2 correspond to this step.
	
	Thus, we have constructed the motion compensation algorithm for manipulator control. At each communication cycle, the final manipulator control algorithm is as described in Algorithm 2.
	
	After applying the motion compensation algorithm, during the actual operation of the manipulator, when there is a significant discrepancy between the desired joint velocity derived from the algorithm and the current velocity of the manipulator, the manipulator is unable to adjust to the desired joint velocity, causing the actual path to deviate from the expected path. Therefore, it is necessary to impose stricter constraints on the joint angular velocity of the desired path to ensure that the manipulator can return to the expected path after deviating. When using dynamic programming to compute redundancy resolution, we modify the velocity upper limit of the joint angles to $\mathbf{\dot{q}}_\text{max2}$.
	
	Subsequently, we will calculate the maximum error introduced by the motion compensation algorithm, as well as the maximum time required for the manipulator to return to the expected path after deviating. We consider the most extreme scenario, where the manipulator and the desired path move in opposite directions at their maximum velocities. Through straightforward calculations, we can determine the maximum joint angle error $\mathbf{err}_\text{max}$ and the maximum time $\mathbf{t}_\text{max}$ required for each joint to return to the desired joint angles:
	\begin{equation}
		\begin{aligned}
			\mathbf{err}_\text{max}=\frac{(\mathbf{\dot{q}}_\text{max}+\mathbf{\dot{q}}_\text{max2})^2}{2\mathbf{\ddot{q}}_\text{max}}\\
			\mathbf{t}_\text{max}=\frac{2{\mathbf{\dot{q}}_\text{max}}^2}{\mathbf{\ddot{q}}_\text{max}(\mathbf{\dot{q}}_\text{max}-\mathbf{\dot{q}}_\text{max2})}
		\end{aligned}		
	\end{equation}
	
	When $\mathbf{\dot{q}}_\text{max2}$ is fixed, $\mathbf{t}_\text{max}$ reaches its minimum value when $\mathbf{\dot{q}}_\text{max}=2\mathbf{\dot{q}}_\text{max2}$. As $\mathbf{\dot{q}}_\text{max2}$ decreases, $\mathbf{err}_\text{max}$ and $\mathbf{t}_\text{max}$ both decreases, while the maximum difference between adjacent adjustment parameters also decreases, which will diminish the algorithm's ability to adjust the path in real time. However, considering the algorithm's ability to adjust the path in real time, it can compensate for errors by dynamically modifying the path in practical applications.

	\section{Results}

	To verify the feasibility of the proposed algorithm, we selected the Franka Emika manipulator for the research and testing of our algorithm. 
	
	\subsection{Comparison with the Franka Cartesian pose generator}
	
	The robot controller has a Cartesian pose generator that can calculate the inverse kinematic solution of the Cartesian pose based on the current state of the manipulator and control its movements accordingly at a fixed frequency. The Cartesian pose generator is an online algorithm, which may lead to interruptions in subsequent motions. We will use the Cartesian path $\mathbf{T}_\text{EE}(t)$ in equation (9) as the test path. 
	
	First, we conducted tests on the proposed algorithm. For the original path, we simulated the joint angles of the manipulator at each communication point using software, obtaining the theoretical results of the algorithm. Additionally, we performed real-world testing using the Franka manipulator, resulting in laboratory results. Whether in simulation or in real-world testing, the manipulator completed the entire motion along the path, and the executed trajectory closely matched the expected path, as is shown in Figure 6.
	
	\begin{figure}
		\centering
		\includegraphics[width=3in]{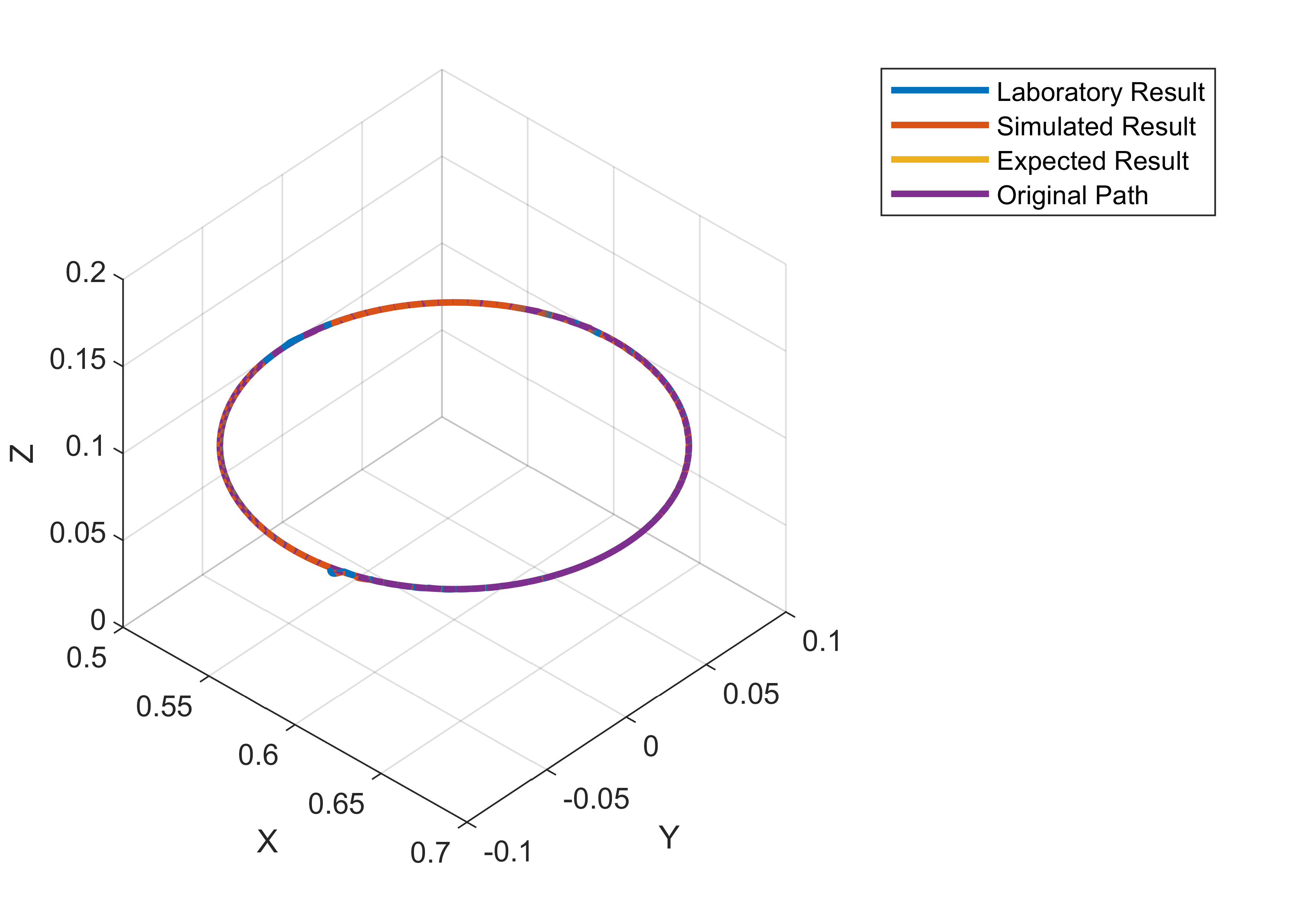}
		\caption{The end-effector paths obtained from both simulated and laboratory tests of the manipulator. Whether in simulation or in real-world testing, the manipulator completed the entire motion along the path, and the executed trajectory closely matched the expected path.} 
		
		\label{fig_5}
	\end{figure}
	
	Subsequently, we computed the angles, velocities, accelerations, and jerks of each joint in the simulated tests, as depicted in Figure 7. All the data has been normalized, with 1 representing its maximum value and -1 representing its minimum value. From the figures, it is evident that all motion parameters adhere to their respective constraints.
	
	\begin{figure}
		\centering
		\includegraphics[width=3.7in]{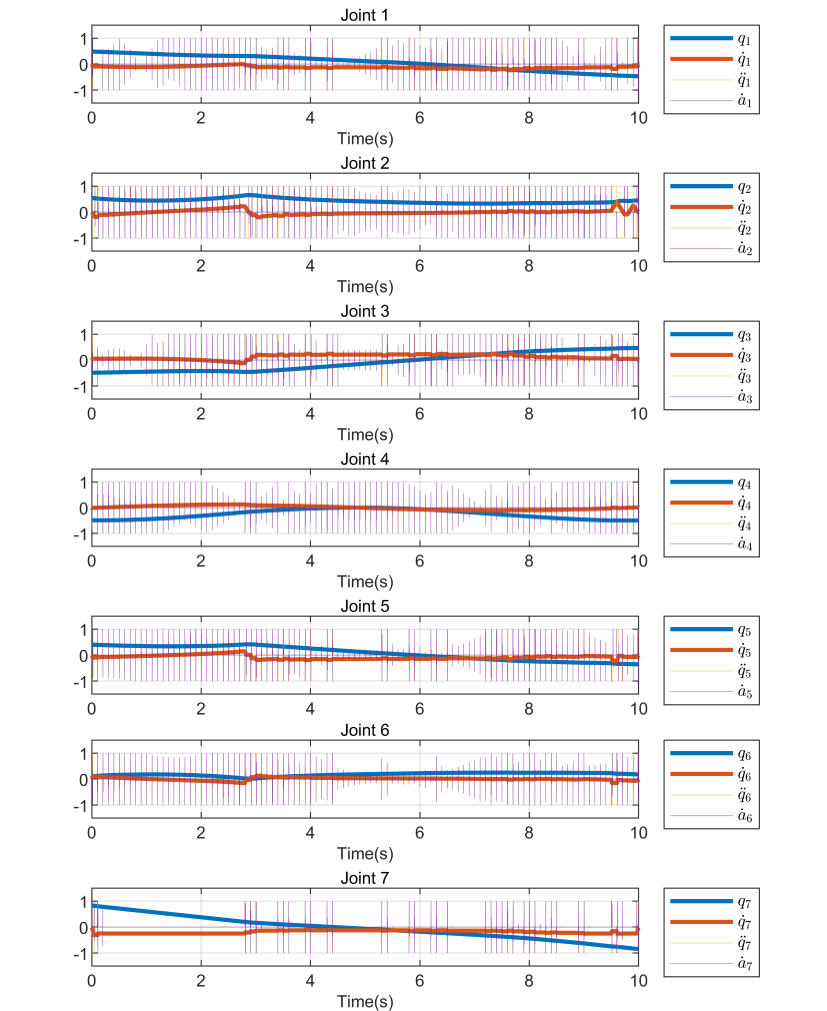}
		\caption{The angles, velocities, accelerations, and jerks of each joint in the simulated tests. All the data has been normalized, with 1 representing its maximum value and -1 representing its minimum value. From the figures, it is evident that all motion parameters adhere to their respective constraints.} 
		
		\label{fig_6}
	\end{figure}
	
	The testing of the Franka Cartesian pose generator, however, did not proceed as smoothly. To maintain control over variables, we initiated the manipulator from the same initial joint angles as the proposed algorithm. Due to joint velocities and accelerations exceeding limits, the manipulator was forced to halt shortly after commencing its motion. Consequently, we had to modify the path velocity to allow for a gradual acceleration of the manipulator. While the manipulator successfully initiated motion, it was compelled to stop due to joint angles surpassing their limits, just after the halfway point of the motion. The end-effector trajectory of the manipulator is illustrated in Figure 8.
	
	\begin{figure}
		\centering
		\includegraphics[width=3in]{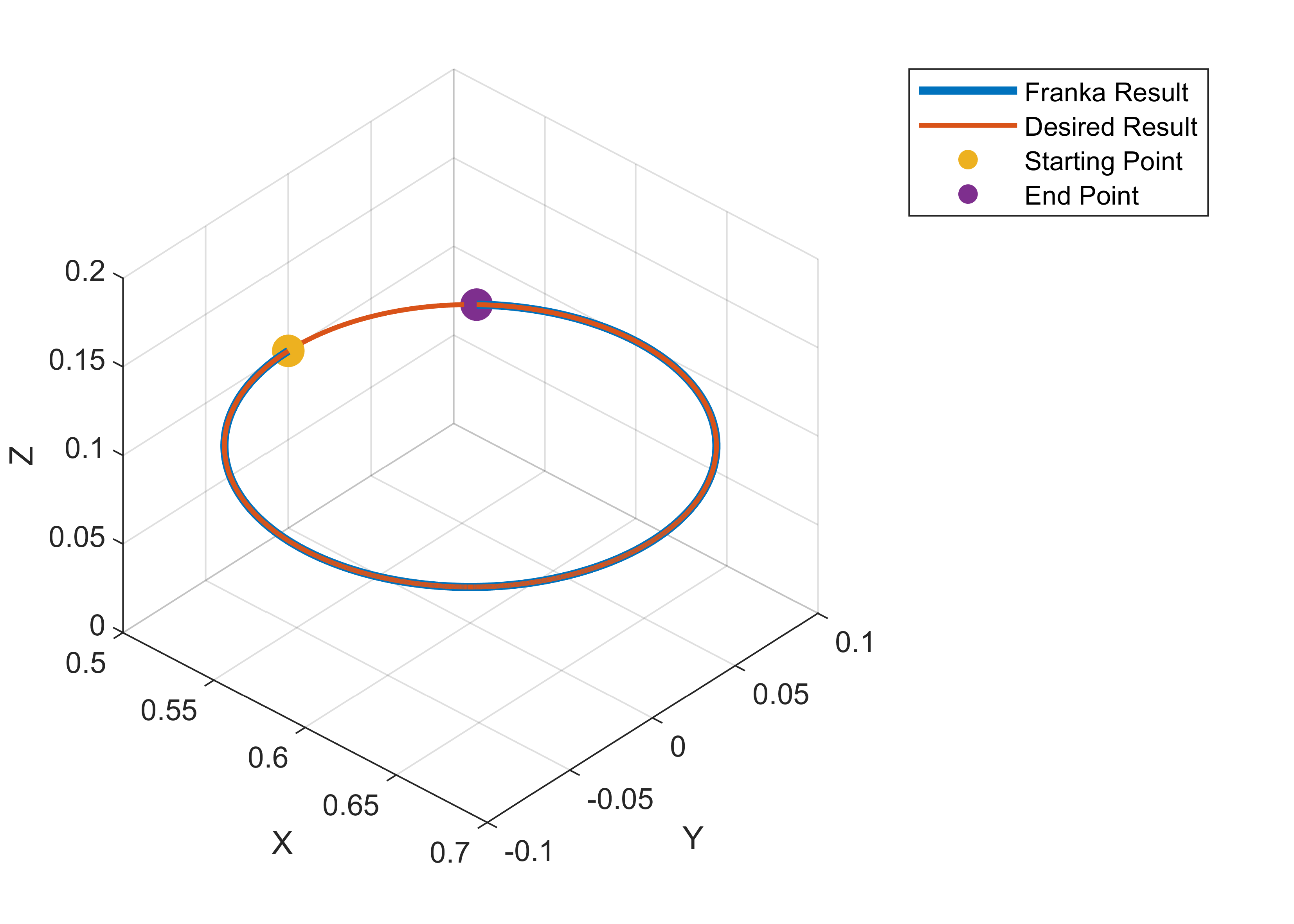}
		\caption{The end-effector trajectory of the manipulator using Franka Cartesian pose generator. The manipulator compelled to stop due to joint angles surpassing their limits.} 
		
		\label{fig_7}
	\end{figure}
	
	Upon inspecting the motion parameters of each joint angle, we discovered that the angle of joint 7 reached its minimum value, resulting in an inability to continue the motion. The angle, velocity, acceleration, and jerk of joint 7 are illustrated in Figure 9, with each datum undergoing the same normalization process as depicted in Figure 7.
	
	\begin{figure}
		\centering
		\includegraphics[width=3.5in]{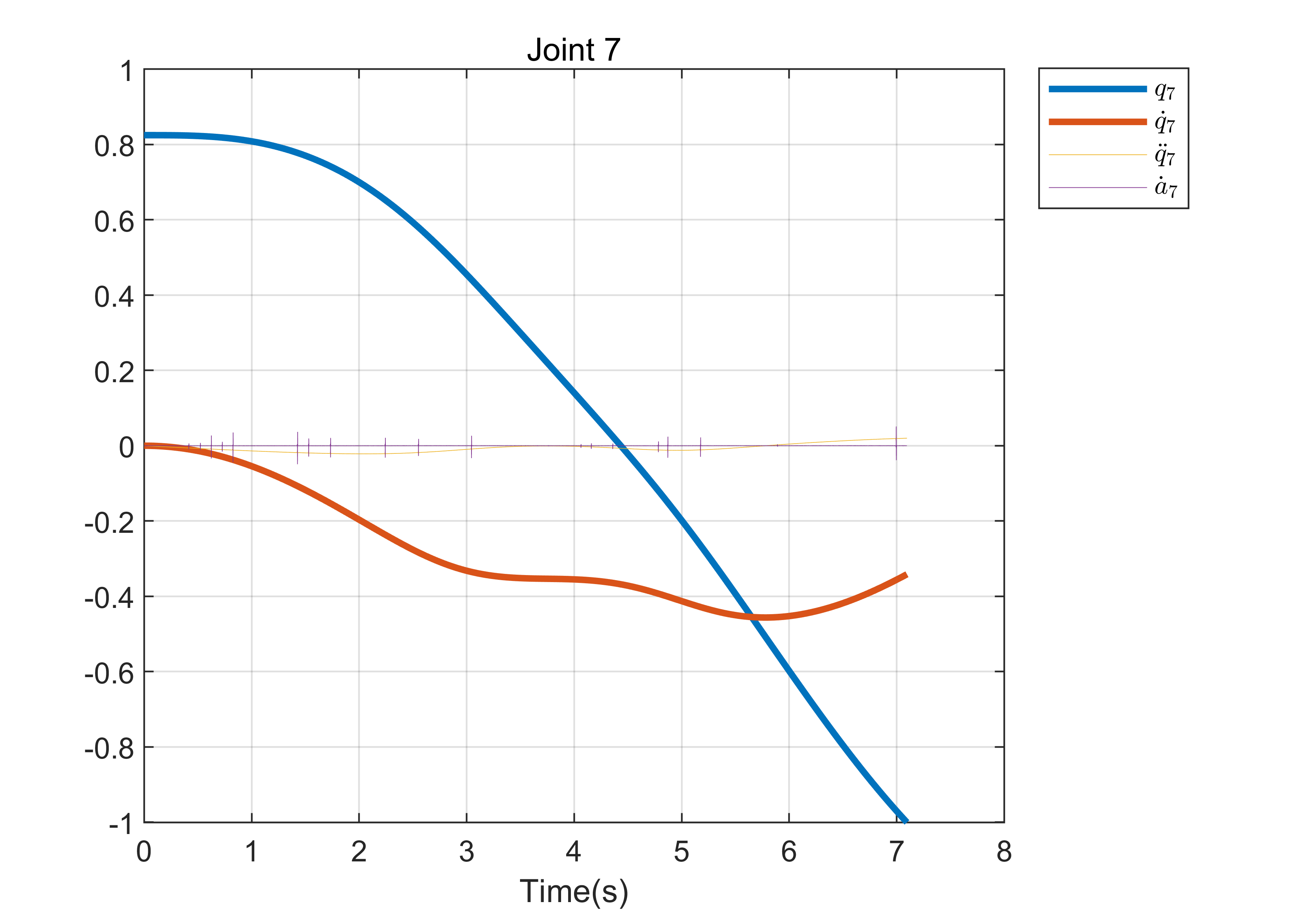}
		\caption{The angle, velocity, acceleration, and jerk of joint 7, with each datum undergoing the same normalization process as depicted in Figure 7. The angle of joint 7 reached its minimum value, resulting in an inability to continue the motion. } 
		
		\label{fig_8}
	\end{figure}
	
	Subsequently, we compared the results obtained from the proposed algorithm with those from the Cartesian pose generator in Figure 10. It is visually apparent that while both methods initiated from the same initial joint angles, as the motion progressed, there were variations in the selection of joint angles. Under the proposed algorithm, the manipulator returns to its original pose, whereas under the Franka algorithm, the manipulator stops midway.
	
	\begin{figure*}[hb]
		\centering
		{\includegraphics[width=7in]{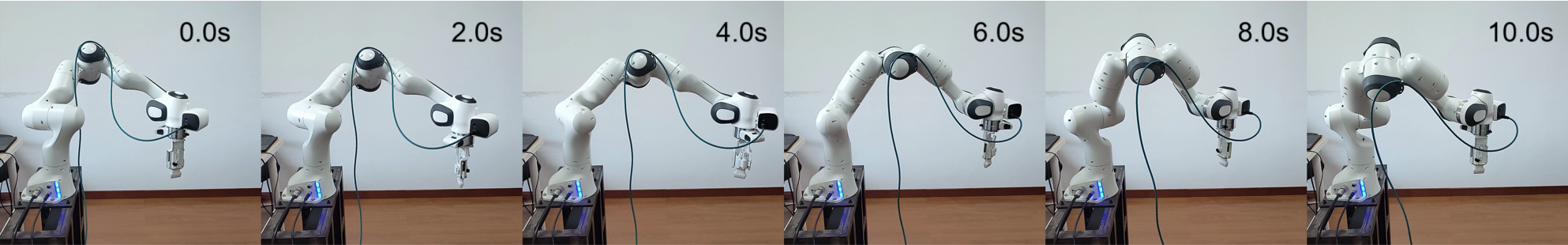}%
			\label{a}}
		\hfil
		{\includegraphics[width=7in]{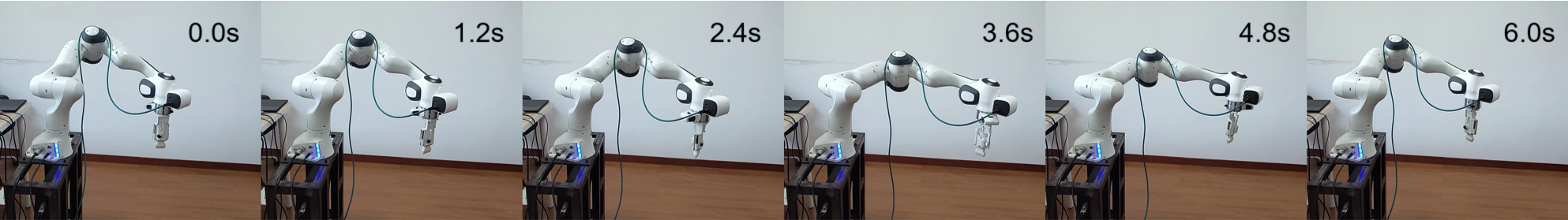}%
			\label{b}}
		\caption{The top row illustrates the results obtained from the proposed algorithm, while the bottom row shows the results of the Franka algorithm. While both methods initiated from the same initial joint angles, as the motion progressed, there were variations in the selection of joint angles. Under the proposed algorithm, the manipulator returns to its original pose, whereas under the Franka algorithm, the manipulator stops midway.}
		\label{fig_9}
	\end{figure*}
	
	By comparison, we observe that the online control algorithm cannot guarantee manipulator completes the entire trajectory. The test path we used is a circle, meaning the manipulator has the same pose at the start and end points. However, by examining Figure 3, we find that the 7th joint angle $q_7$ of the manipulator must decrease from its maximum to its minimum value in order to ensure movement along the path. While the start and end points have the same pose, different joint angles are required considering the entire trajectory, which the online control algorithm is unable to achieve. This comparative experiment highlights the advantages of the offline planning algorithm.

	\subsection{Practical task test}
	
	Next, we tested the real-time adjustment capability of the proposed algorithm with a practical task. This is the key highlight of the proposed algorithm, a capability that other offline planning algorithms are unable to achieve. A model probe is mounted in a sliding groove at the end of the manipulator and is connected to the manipulator via a spring, allowing it to make small movements along the direction of the manipulator's end-effector. Subsequently, a distance sensor was attached to the end-effector to measure the distance between the probe and the end-effector. 
	
	During the manipulator's motion, the probe was directly contacted by hand. By adjusting the path in real-time based on the sensor readings and keeping the readings within a fixed range, the manipulator was able to move while simultaneously following the hand's vertical adjustments. The test path with 101 sampling points is defined as follows:
	
	\begin{equation}
		\mathbf{{T}}_\text{EEi} = \begin{bmatrix}
			\cos(\frac{2\pi i}{100})	& \sin(\frac{2\pi i}{100}) & 0 &0.6+0.1\cos(\frac{2\pi i}{100}) \\
			\sin(\frac{2\pi i}{100})	& -\cos(\frac{2\pi i}{100}) & 0 & 0.1\sin(\frac{2\pi i}{100})\\
			0	& 0 & -1 & 0.2\\
			0	& 0 &0  &1
		\end{bmatrix}
	\end{equation}
	The time interval between two adjacent sampling points is 0.1s. The maximum adjustment parameter $b_o$ is 0.05m. Figure 11 illustrates the testing of the manipulator's path adjustment in response to hand movements. Before the hand makes contact with the manipulator, the manipulator actively adjusts its trajectory downward along the direction of the end-effector, effectively performing a target-seeking action. Once the hand contacts the model probe, the manipulator can actively adjust its motion trajectory in real-time to follow the vertical movements of the hand.
	
	\begin{figure*}[h]
		\centering
		
		\includegraphics[width=6in]{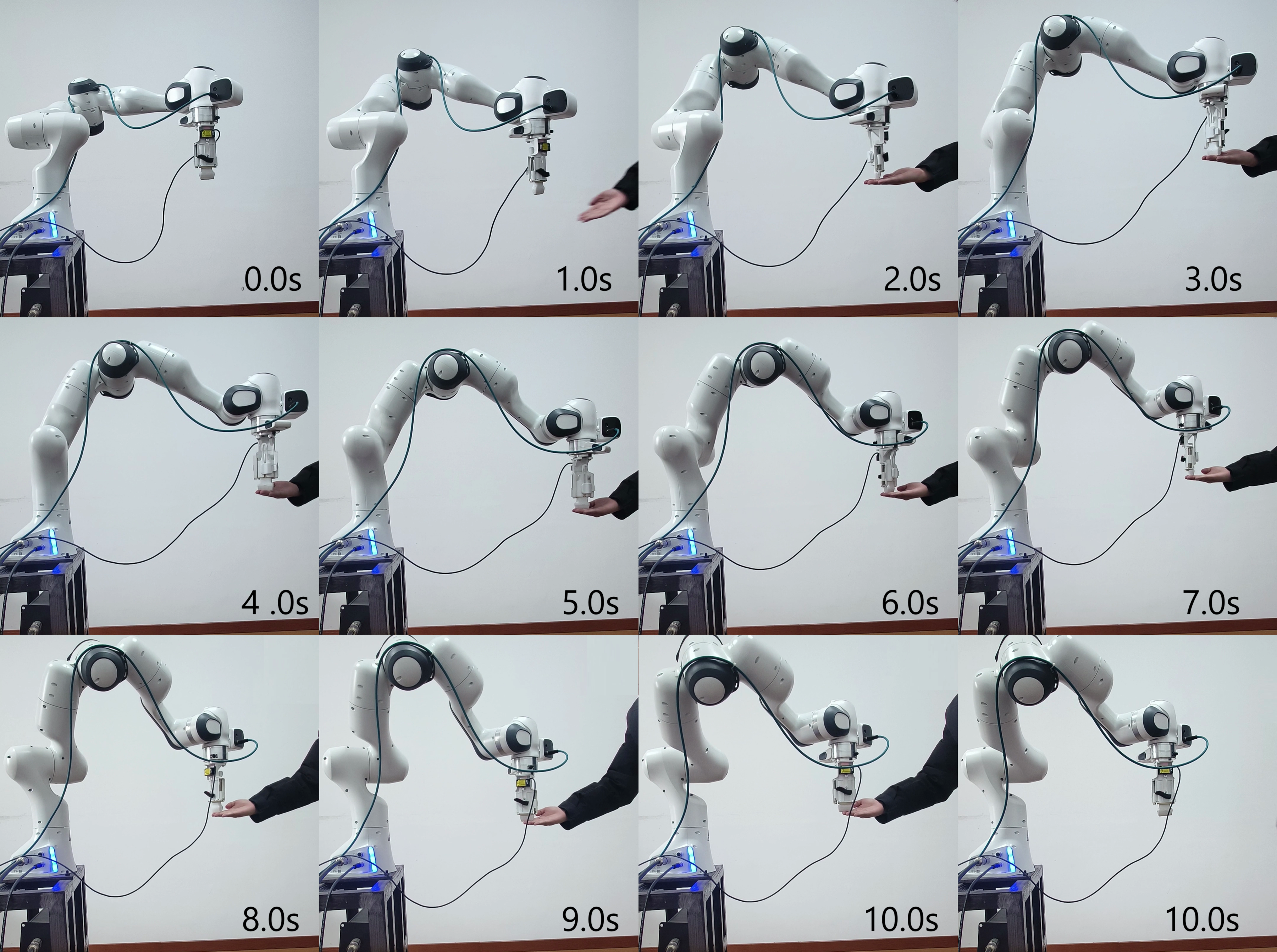}%
		
		\caption{The testing of the manipulator's path adjustment in response to hand movements. Before the hand makes contact with the manipulator, the manipulator actively adjusts its trajectory downward along the direction of the end-effector, effectively performing a target-seeking action. Once the hand contacts the model probe, the manipulator can actively adjust its motion trajectory in real-time to follow the vertical movements of the hand}
		\label{fig_14}
	\end{figure*}
	
	Then we plotted the trajectory of the manipulator's end-effector in Figure 12. The original trajectory of the manipulator, the desired adjusted trajectory computed in real-time by the proposed algorithm, and the actual trajectory of the manipulator after motion compensation were all included in the plot. It can be observed that the manipulator effectively follows the adjusted trajectory, and its path exhibits real-time adjustments compared to the original trajectory. These results demonstrate the feasibility and reliability of the proposed algorithm.
	
	\begin{figure}
		\centering
		\includegraphics[width=3in]{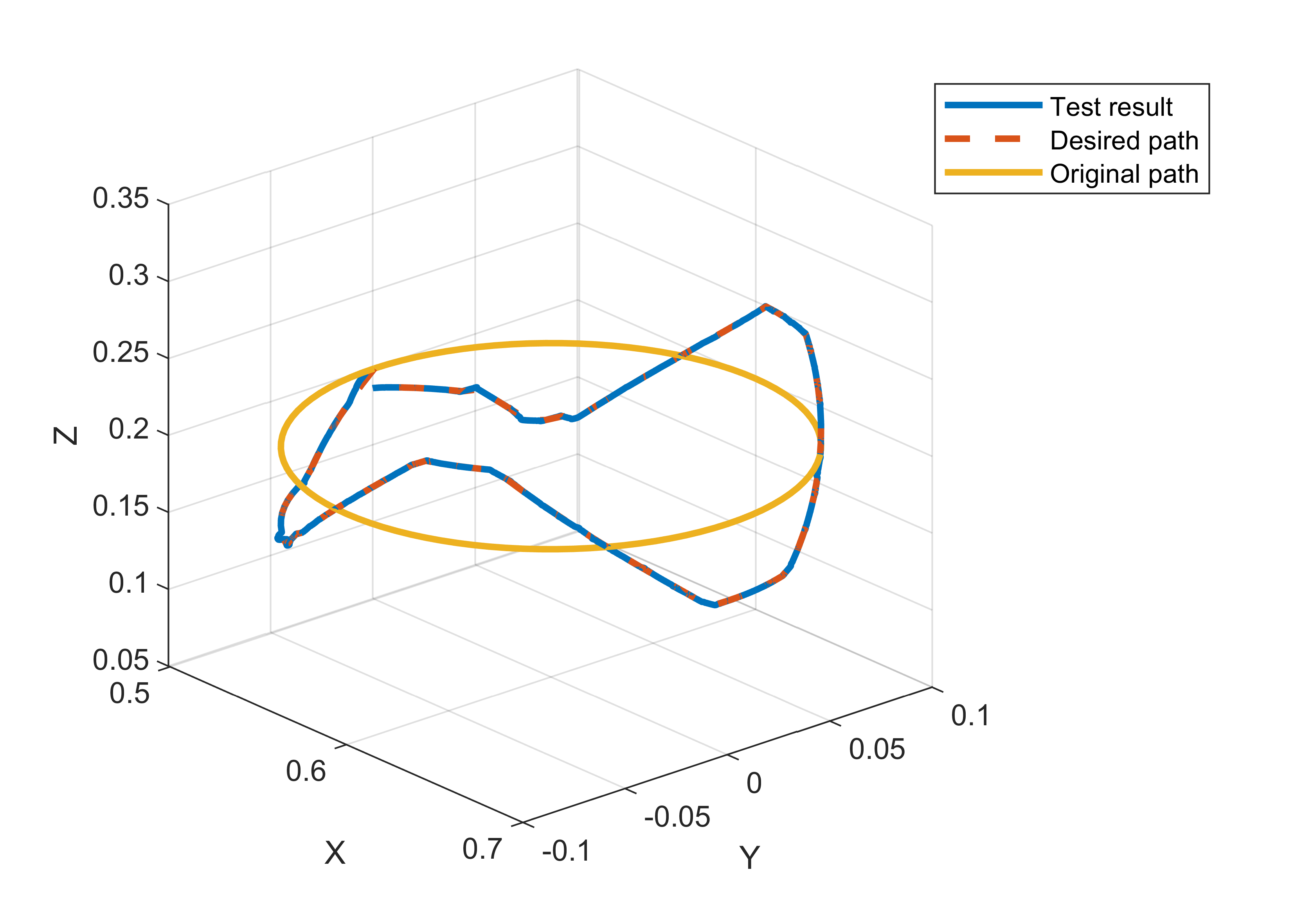}
		\caption{The original trajectory of the manipulator, the desired adjusted trajectory computed in real-time by the proposed algorithm, and the actual trajectory of the manipulator after motion compensation were all included in the plot. It can be observed that the manipulator effectively follows the adjusted trajectory, and its path exhibits real-time adjustments compared to the original trajectory.} 
		
		\label{fig_15}
	\end{figure}
	
	Subsequently, we computed the error in the results. To intuitively demonstrate the error, we only considered the positional error along the path. During the operation of the manipulator, while controlling the manipulator to move along the path in real time, we also calculated the desired path after motion compensation. In Figure 13, we plot the error of the calculated results and the actual experimental results compared to the desired path obtained by the algorithm. Larger errors occur at the starting and ending points of the path due to the acceleration from zero velocity and deceleration to zero velocity. The errors for both results are maintained within the order of $10^{-3}$ meters. The error in the actual experimental results is higher than that in the calculated results due to the manipulator's control.
	
	\begin{figure}
		\centering
		\includegraphics[width=3in]{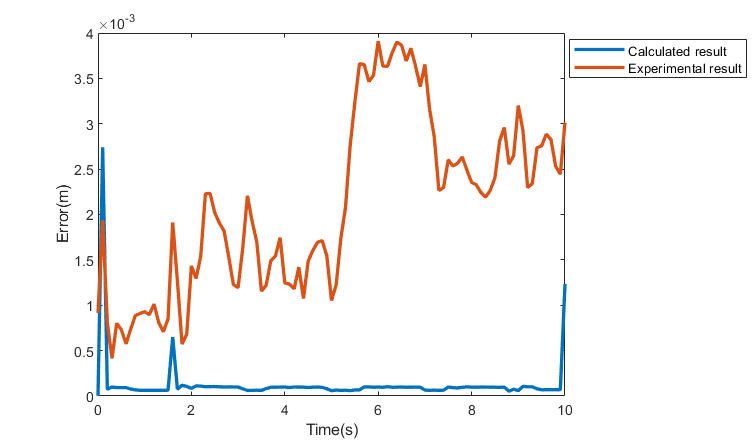}
		\caption{The error of the calculated results and the actual experimental results compared to the desired path obtained by the algorithm. Larger errors occur at the starting and ending points of the path due to the acceleration from zero velocity and deceleration to zero velocity. The errors for both results are maintained within the order of $10^{-3}$ meters. The error in the actual experimental results is higher than that in the calculated results due to the manipulator's control.} 
		
		\label{fig_16}
	\end{figure}
	
	From the experimental results, it can be observed that the proposed algorithm effectively enables real-time adjustment of the manipulator's path. The algorithm itself incurs negligible computational errors, and the errors generated by the motion compensation interpolation algorithm corresponding to the proposed algorithm are also within the fine-tuning range of devices such as springs. Therefore, the proposed algorithm can effectively accomplish its intended function.
	
	\section{Conclusion}
	
	This paper innovatively proposes a dynamic programming-based offline redundancy resolution of redundant manipulators along prescribed paths with real-time adjustment. The proposed algorithm not only retains the advantages of offline planning by guaranteeing the feasibility of the manipulator's motion but also enables real-time adjustment of the subsequent path based on feedback from sensors and other sources during operation, thereby achieving online control. The proposed algorithm represents a significant breakthrough in offline planning algorithms for manipulators. 
	
	However, due to the uncertainty of the path generated by the algorithm, it is not possible to apply smoothing techniques such as interpolation to the path before the manipulator's operation. Instead, motion compensation is employed to minimize the discrepancy between the actual and desired paths as much as possible. When the manipulator's joints undergo frequent acceleration, deceleration, or direction changes, errors and vibrations may occur. This also results in the algorithm performing not as well for complex paths. Therefore, in future research, we aim to improve the smoothness of the path to enhance the algorithm's performance.

	\section*{Acknowledgments}
	This work was supported by The National Natural Science Foundation(Nos.12090020, 12090025) and the Zhejiang Provincial Science and Technology Program(2022C03113).
	
	\printcredits
	
	\bibliographystyle{cas-model2-names}

	\bibliography{cas-refs}
	
\end{document}